%% file: main.tex
\documentclass[letterpaper]{article}
\usepackage[preprint]{aaai2027}
\usepackage[hyphens]{url}  
\usepackage{graphicx} 
\usepackage{natbib}  
\usepackage{caption} 
\usepackage{amsfonts} 
\usepackage{amsmath} 
\usepackage{booktabs} 
\usepackage{multirow} 
\usepackage{algorithm}
\usepackage{algorithmic}

\title{LAB-Tab: LLM-Augmented Bayesian Network Adaptation for Few-Shot Tabular Generation}

\author{
    Zijian Shen,
    Taijie Chen,
    Bin Zhou,
    Ziyang Jiang,
    Jintao Ke\corresponding
}

\affiliations{
    Department of Civil Engineering, The University of Hong Kong
}

\begin{document}

\maketitle

\begin{abstract}
Tabular data generation supports analysis and decision-making when target-domain data are scarce, yet collecting complete target samples is often costly. A practical but underexplored setting provides only a few target records together with richer source data from a related domain. Existing few-shot tabular generators often either fit sparse target statistics directly, which can overfit incidental patterns, or reuse source-domain generators, which may preserve dependencies that no longer hold in the target domain. To address this problem, we propose \textbf{LAB-Tab}, an LLM-augmented Bayesian network (BN) adaptation framework for source-aware few-shot tabular generation. LAB-Tab first fits a BN from source data and then uses an LLM to propose plausible target-domain BN edges that are absent from the source BN graph. This step converts semantic and weak statistical evidence into explicit structural hypotheses, thereby expanding the editable edge space beyond the source-fitted graph.  Because the proposed edges may be noisy and interact with existing dependencies, a PPO policy calibrates edges in the augmented BN through edge-level actions, including \textit{keep}, \textit{weaken}, \textit{strengthen}, \textit{flip}, and \textit{deactivate}. The PPO policy is trained with a reward that combines distributional alignment, downstream utility, and preservation of target-relevant dependencies. The adapted BN is then sampled to synthesize target-domain tables. Across six source--target distribution-shift scenarios built from three US Census (ACS) prediction tasks, \textbf{LAB-Tab} achieves the best performance at the 10\% target-data budget, leads four of the six individual scenarios, and reduces the macro Overall score by 33.8\% relative to the strongest baseline. It also obtains the best macro JSD, WAPE, and UtilityGap while maintaining competitive feature--label preservation.
\end{abstract}

\section{Introduction}
\begin{figure*}[t]
    \centering
    \includegraphics[width=0.8\textwidth]{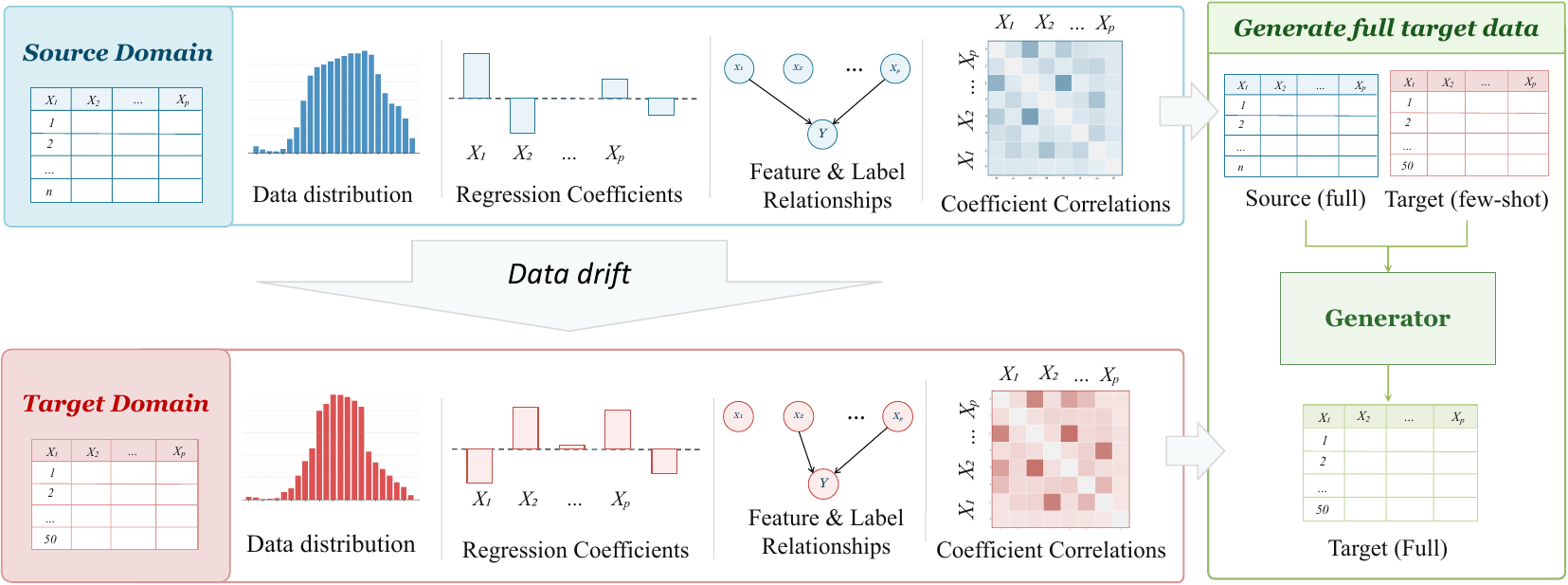}
    \caption{Source-aware few-shot tabular generation under distribution
    shift. A large source table and a small target subset are used to generate
    a full synthetic target table. The target domain may differ from the source
    in marginal distributions, predictive coefficients, feature--label
    relationships, and cross-feature associations.}
    \label{fig:intro}
\end{figure*}
Tabular data generation supports data analysis, privacy-preserving sharing, simulation, and benchmark construction across domains such as healthcare, finance, transportation, and public-sector analytics. In many real-world applications, however, obtaining a representative target-domain table is difficult. Collecting sufficient target-domain records, together with the annotations required for downstream tasks, is often costly and time-consuming \citep{zhuang2021comprehensive}. Even when some records are accessible, they may be sparse, incomplete, or affected by missing values \citep{dong2013principled,emmanuel2021survey}. Consequently, practitioners may have access to only a small number of target-domain records. In this low-data regime, both marginal distributions and multivariate dependencies must be estimated under substantial uncertainty, making it difficult to learn a target-domain generator without overfitting incidental patterns. This practical challenge motivates the problem of few-shot tabular generation.

Although target-domain data are scarce in this setting, a substantially larger table from a related source domain is often available. The source table may originate from an earlier survey, another geographic region, or a broader population, and can provide more reliable statistical information than the few observed target records. However, source-domain information cannot be transferred indiscriminately. As illustrated in Figure~\ref{fig:intro}, distribution shift may affect both column marginals and conditional relationships among variables \citep{quinonero2009dataset,koh2021wilds}. Some dependencies from the source data may remain useful, while others may change dramatically within the target-specific relationships. For example, education may remain associated with income across regions, whereas the relationship between working hours and income may vary with local labor markets. The central challenge is therefore to exploit reliable source-domain knowledge while accommodating target-specific changes supported by only limited evidence.

Existing tabular generators, transfer-learning methods, and recent LLM-based approaches offer complementary capabilities in distribution modeling, cross-domain reuse, and semantic guidance \citep{solatorio2023realtabformer,shi2025survey,yang2025spada,chen2026addressing}. However, they do not jointly support explicit source-to-target dependency adaptation and feedback-driven calibration under limited target evidence. To address this gap, we propose \textbf{LAB-Tab}, an LLM-augmented Bayesian network (BN) adaptation framework for source-aware few-shot tabular generation.  LAB-Tab first learns a BN from the source table, using its explicit structure and conditional distributions as a stable and inspectable prior. Then, LAB-Tab utilizes an LLM to propose target-relevant edges absent from the source graph, expanding the structural search space through sparse target data. Finally, PPO calibrates these dependencies using target-domain data, resolving inaccurate edges before sampling. This design allows transferable source dependencies to be retained while introducing structural changes based on few-shot target-specific dataset.

Our contributions are summarized as follows:

\begin{itemize}
\item We introduce \textbf{LAB-Tab}, a source-aware few-shot tabular generation framework that combines a source-fitted BN, LLM-guided candidate-edge augmentation, and PPO-based dependency calibration.

\item We formulate source-to-target adaptation as explicit BN edge control. The LLM expands the source-derived structural search space, deterministic validation enforces graph validity, and PPO adjusts edge activation, magnitude, and polarity using complementary global table-level and local dependency-level feedback.

\item We evaluate LAB-Tab on six source--target shifts derived from three 2018
American Community Survey (ACS) prediction tasks. At the 10\% target-data budget,
LAB-Tab achieves the lowest macro Overall score, leads four of the six
scenarios, and obtains the best macro JSD, WAPE, and UtilityGap. The ablations
indicate complementary contributions from candidate-edge augmentation, PPO
control, and local dependency feedback within the evaluated design.
\end{itemize}

\section{Related Work}

Tabular synthesis spans explicit probabilistic and high-capacity neural generators \citep{shi2025survey,challagundla2025survey,stoian2025survey}. Bayesian-network, SDV-style, and copula methods offer efficient sampling with inspectable dependency assumptions \citep{zhang2014privbayes,patki2016sdv,asghar2019dpgaussiancopula}, whereas GAN and diffusion models increase expressiveness \citep{xu2019ctgan,kotelnikov2023tabddpm,zhang2024tabsyn}. Most remain target-table driven. With few target records, neural models may fit unstable patterns, while probabilistic models must estimate local conditionals from sparse counts. When a related source table is available, the problem becomes generative domain adaptation. Conventional transfer learning improves target prediction through representation transfer, reweighting, distribution matching, or fine-tuning, but does not explicitly determine which dependencies a target generator should retain, introduce, or recalibrate \citep{zhuang2021comprehensive}.

LLM-based methods add schema semantics and in-context reasoning to tabular synthesis. GReaT, REaLTabFormer, and Tabula generate records from serialized rows or relations \citep{borisov2023great,solatorio2023realtabformer,zhao2023tabula}, while subsequent methods improve example selection, efficiency, logical consistency, or dependency guidance \citep{xu2024llmsnatural,fang2025tabgenic,nguyen2025fastgen,long2025llmtablogic}. In particular, SPADA uses an LLM-induced dependency graph to guide synthesis \citep{yang2025spada}. Feedback-based learning and LLM-guided search further demonstrate the value of external evaluation for refining model outputs or proposed solutions \citep{christiano2017deep,ouyang2022training,romera2024funsearch,ma2024eureka}. Taken together, prior work largely treats semantic structure induction, probabilistic generation, and feedback-driven adaptation as separate directions. Our proposed method bridges this gap by unifying these complementary capabilities in a framework for source-aware few-shot tabular generation.

\section{Problem Formulation}

Let $\mathcal{X}=\{X_1,\ldots,X_p\}$ denote the features, $Y$ the task label, and $\mathcal{V}=\mathcal{X}\cup\{Y\}=\{V_1,\ldots,V_{p+1}\}$ the shared source--target schema. A record is $\mathbf{z}=(\mathbf{x},y)$. We observe an abundant source table $\mathcal{D}_s=\{\mathbf{z}_i^s\}_{i=1}^{n_s}$ drawn i.i.d.\ from $P_s$ and a few-shot target table $\mathcal{D}_t^{\mathrm{fs}}=\{\mathbf{z}_i^t\}_{i=1}^{K}$ drawn i.i.d.\ from $P_t$, where $K\ll n_s$. The two distributions may differ in both marginals and conditional dependencies. During adaptation, $P_t$ is unknown and held-out target records are used only for evaluation. Let $\mathcal{B}(\mathcal{V})$ denote the space of BN generators over $\mathcal{V}$. A source-aware adaptation rule produces
\begin{equation}
\begin{aligned}
B^\star
&=
\mathcal{A}
\left(
\mathcal{D}_s,
\mathcal{D}_t^{\mathrm{fs}}
\right)
\in\mathcal{B}(\mathcal{V}),\\
\widehat{\mathcal{D}}_t
&=
\{\widehat{\mathbf{z}}_i\}_{i=1}^{N},
\qquad
\widehat{\mathbf{z}}_i
\overset{\mathrm{i.i.d.}}{\sim}
P_{B^\star}.
\end{aligned}
\end{equation}

A BN is $B=(\mathcal{G},\Theta)$, where $\mathcal{G}=(\mathcal{V},\mathcal{E})$ is a directed acyclic graph and $\Theta=\{\theta_j\}_{j=1}^{p+1}$ contains its local conditional distributions. For $V_j\in\mathcal{V}$, let $\mathrm{Pa}_{\mathcal{G}}(V_j)=\{V_i:(V_i,V_j)\in\mathcal{E}\}$ and
$\theta_j(v_j\mid\mathbf{pa}_j)=P_B(V_j=v_j\mid\mathrm{Pa}_{\mathcal{G}}(V_j)=\mathbf{pa}_j)$. The induced joint distribution factorizes as
\begin{equation}
P_B(\mathbf{z})
=
\prod_{j=1}^{p+1}
\theta_j(v_j\mid\mathbf{pa}_j).
\end{equation}

The target-aligned generator should balance distributional fidelity, predictive utility, and feature--label preservation. Let $D_{\mathrm{dist}}(P_B,P_t)\geq0$ denote distributional discrepancy. For a fixed training size $n_u$, let $U_{n_u}(Q;P_t)$ denote the expected target performance of a fixed learner trained on samples from $Q$, and let $\mathbf{a}(Q)$ be the coefficient vector of a fixed linear probe under $Q$. Define
\begin{equation}
\begin{aligned}
\Delta_U(B)
&=
\left|
U_{n_u}(P_B;P_t)
-
U_{n_u}(P_t;P_t)
\right|,\\
\Delta_A(B)
&=
1-
\frac{
\mathbf{a}(P_B)^\top\mathbf{a}(P_t)
}{
\|\mathbf{a}(P_B)\|_2
\|\mathbf{a}(P_t)\|_2
}.
\end{aligned}
\end{equation}
The population objective and its ideal solution are
\begin{equation}
\begin{aligned}
\mathcal{L}_t(B)
&=
\lambda_{\mathrm{d}}D_{\mathrm{dist}}(P_B,P_t)
+
\lambda_{\mathrm{u}}\Delta_U(B)
+
\lambda_{\mathrm{a}}\Delta_A(B),\\
B^\dagger
&\in
\arg\min_{B\in\mathcal{B}(\mathcal{V})}
\mathcal{L}_t(B),
\end{aligned}
\end{equation}
where the nonnegative weights are not all zero. Since $P_t$ is unknown, the practical objective is to construct $B^\star$ from $\mathcal{D}_s$ and $\mathcal{D}_t^{\mathrm{fs}}$ whose induced distribution approximates $P_t$ under these three criteria.

\section{Methodology}

\begin{figure*}[t]
    \centering
    \includegraphics[width=1\textwidth]{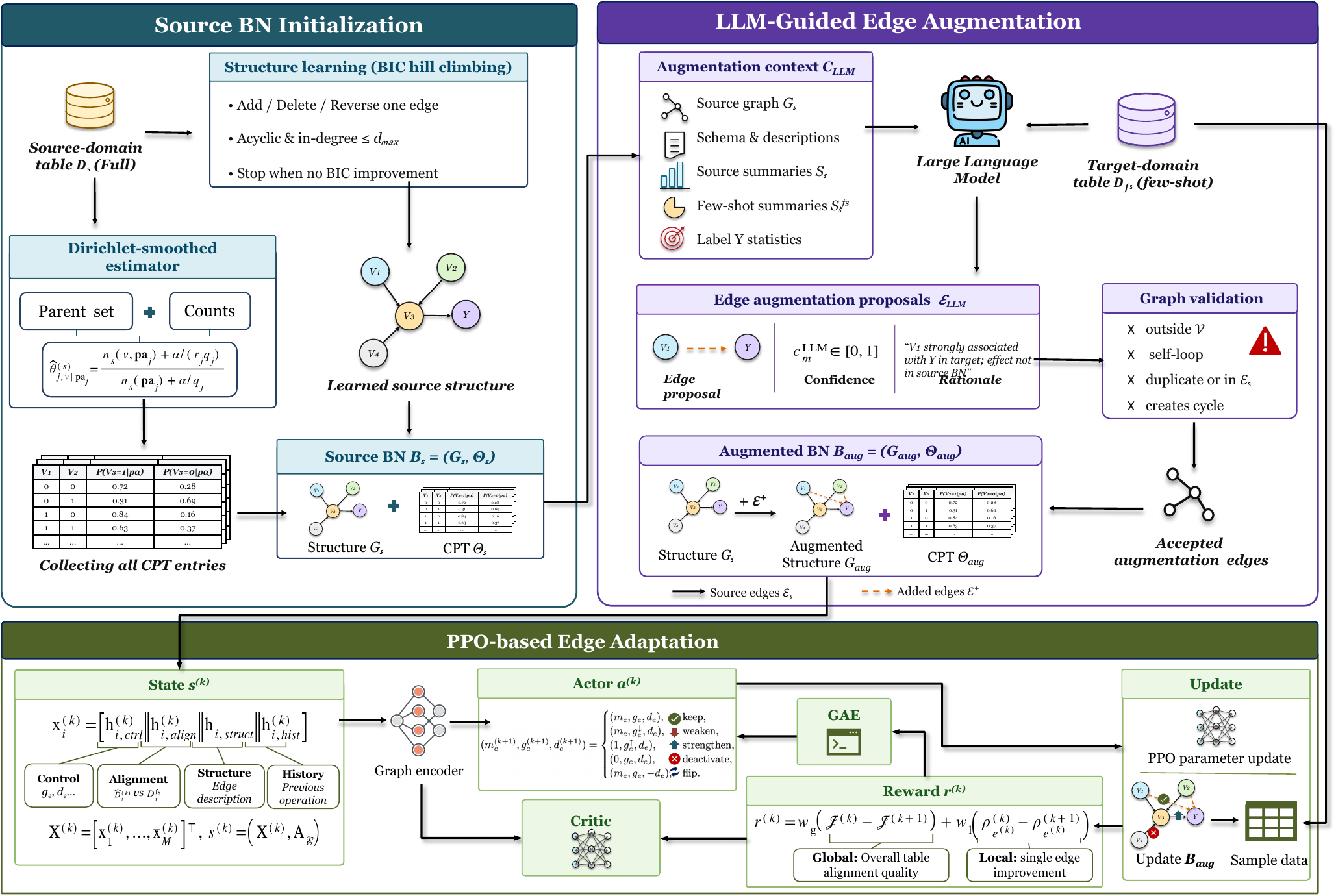}
    \caption{Overview of LAB-Tab: source BN learning and Conditional Probability Table (CPT) estimation, LLM-guided edge augmentation with graph validation, and PPO-based edge
control with global--local rewards.}
    \label{fig:labtab_overview}
\end{figure*}

Figure~\ref{fig:labtab_overview} summarizes LAB-Tab, which adapts a source-fitted BN using the few-shot target set $\mathcal{D}_t^{\mathrm{fs}}$. The framework consists of source BN
initialization, LLM-guided edge augmentation, and PPO-guided edge-control
optimization. The adapted BN is then ancestrally sampled to generate the synthetic target
table. The label variable is denoted by $Y\in\mathcal{V}$ throughout.

\subsection{Source BN Initialization}

We first learn a source BN
$B_s=(\mathcal{G}_s,\Theta_s)$ from the source-domain table
$\mathcal{D}_s$. Each variable is represented by a finite state space, with
continuous variables discretized before structure and parameter learning.

\paragraph{Structure learning.}
Let $\mathcal{G}_s=(\mathcal{V},\mathcal{E}_s)$ denote the source DAG.
Starting from an empty graph, we perform greedy hill climbing under the
Bayesian Information Criterion (BIC). At each step, the search considers
single-edge additions, deletions, and reversals, and applies the valid
operation that yields the largest positive BIC improvement. A candidate
operation is valid only if it preserves acyclicity and keeps the in-degree of
every node below $d_{\max}$. The search terminates when no valid operation
improves the score. Since hill climbing is local,
\begin{equation}
\mathcal{G}_s
\approx
\arg\max_{\mathcal{G}\in\mathrm{DAG}_{d_{\max}}(\mathcal{V})}
\mathrm{BIC}(\mathcal{G};\mathcal{D}_s),
\end{equation}
where $\mathrm{DAG}_{d_{\max}}(\mathcal{V})$ contains DAGs over $\mathcal{V}$ with maximum in-degree $d_{\max}$.

\paragraph{Parameter estimation.}
For each variable $V_j$, let
$\mathcal{P}_j^s=\mathrm{Pa}_{\mathcal{G}_s}(V_j)$ denote its parent set,
$r_j$ the number of states of $V_j$, and $q_j$ the number of joint
configurations of $\mathcal{P}_j^s$. For a state
$v\in\mathrm{Val}(V_j)$ and a parent configuration
$\mathbf{pa}_j\in\mathrm{Val}(\mathcal{P}_j^s)$, we estimate the source CPT
using a symmetric Dirichlet prior:
\begin{equation}
\widehat{\theta}^{(s)}_{j,v\mid\mathbf{pa}_j}
=
\frac{
n_s(v,\mathbf{pa}_j)+\alpha/(r_jq_j)
}{
n_s(\mathbf{pa}_j)+\alpha/q_j
},
\end{equation}
where $n_s(v,\mathbf{pa}_j)$ is the number of source records satisfying
$V_j=v$ and $\mathcal{P}_j^s=\mathbf{pa}_j$, and
$n_s(\mathbf{pa}_j)=\sum_v n_s(v,\mathbf{pa}_j)$. The hyperparameter
$\alpha$ denotes the equivalent sample size. Collecting the CPTs of all
variables gives $\Theta_s$ and hence the source BN is derived as $B_s=(\mathcal{G}_s,\Theta_s)$.

\subsection{LLM-Guided BN Edge Augmentation}

The source BN captures dependencies supported by $\mathcal{D}_s$, but
target-relevant dependencies may be absent from $\mathcal{G}_s$ under domain
shift. Relearning the entire DAG from the few-shot target set
$\mathcal{D}_t^{\mathrm{fs}}$ is unreliable because many candidate relations
receive limited empirical support. LAB-Tab therefore retains the source graph
as a structural anchor and uses an LLM to propose a small set of candidate edge
additions.

\paragraph{LLM-guided edge proposal.}
The augmentation context $\mathcal{C}_{\mathrm{LLM}}$ contains the source graph
$\mathcal{G}_s$, the shared schema and variable descriptions, source-domain
summaries $\mathcal{S}_s$, and few-shot target summaries
$\mathcal{S}_t^{\mathrm{fs}}$. The summaries include category support,
marginal frequencies, pairwise associations, and label-conditional contrasts.
Given this context, the LLM returns
\begin{equation}
\widetilde{\mathcal{P}}_{\mathrm{LLM}}
=
\Phi_{\mathrm{LLM}}(\mathcal{C}_{\mathrm{LLM}})
=
\left\{
p_m=
\left(
\widetilde e_m,
c_m^{\mathrm{LLM}},
\ell_m
\right)
\right\}_{m=1}^{L},
\end{equation}
where $\widetilde e_m=(V_{u_m}\rightarrow V_{v_m})$ is a proposed directed
edge, $c_m^{\mathrm{LLM}}\in[0,1]$ is a self-assessed confidence score, and
$\ell_m$ is a concise semantic and statistical rationale. The confidence score
and rationale are retained as proposal metadata for subsequent calibration and
auditability. 

\paragraph{Graph validation.}
The proposed edges are processed by
\begin{equation}
\mathcal{E}^{+}
=
\mathrm{ValidAdd}
\left(
\{\widetilde e_m\}_{m=1}^{L},
\mathcal{G}_s
\right).
\end{equation}
The validator rejects edges with endpoints outside $\mathcal{V}$, self-loops,
duplicate or existing edges, and additions that would create a directed cycle
or cause a node's in-degree to exceed $d_{\max}$. Each candidate is checked
against the graph containing all previously accepted additions. The resulting
augmented graph is
\begin{equation}
\begin{aligned}
\mathcal{E}_{\mathrm{aug}}
&=
\mathcal{E}_s\cup\mathcal{E}^{+},\\
\mathcal{G}_{\mathrm{aug}}
&=
(\mathcal{V},\mathcal{E}_{\mathrm{aug}}).
\end{aligned}
\end{equation}

Because the accepted additions alter the parent sets of their child variables,
we re-estimate the conditional distributions from $\mathcal{D}_s$ under
$\mathcal{G}_{\mathrm{aug}}$ using the same Dirichlet estimator. Let
$\bar{\theta}^{(s)}_{j,v\mid\mathbf{pa}_j}$ denote the resulting source
conditional probability. We then combine this source estimate with the
few-shot target counts:
\begin{equation}
\theta^{\mathrm{aug}}_{j,v\mid\mathbf{pa}_j}
=
\frac{
n_t^{\mathrm{fs}}(v,\mathbf{pa}_j)
+
\kappa\bar{\theta}^{(s)}_{j,v\mid\mathbf{pa}_j}
}{
n_t^{\mathrm{fs}}(\mathbf{pa}_j)+\kappa
},
\end{equation}
where $\kappa>0$ is the source-informed prior concentration. Equivalently,
$\bar{\theta}^{(s)}_{j,\cdot\mid\mathbf{pa}_j}$ specifies the prior mean and
$\kappa$ controls its strength relative to the few-shot target observations.
Collecting these entries gives $\Theta_{\mathrm{aug}}$ and the augmented BN
\begin{equation}
B^{\mathrm{aug}}
=
(\mathcal{G}_{\mathrm{aug}},\Theta_{\mathrm{aug}}),
\end{equation}
which is subsequently calibrated by the edge-control policy.

\subsection{PPO-Guided Edge-Control Optimization}

The augmented BN favors structural coverage over exact target calibration.
We therefore formulate edge control as a finite-horizon Markov decision
process and optimize it with PPO~\citep{schulman2017ppo}. For each edge
$e\in\mathcal{E}_{\mathrm{aug}}$, let $m_e\in\{0,1\}$,
$g_e\in[g_{\min},g_{\max}]$, and $d_e\in\{-1,+1\}$ denote its activation,
magnitude, and polarity, respectively, where
$0<g_{\min}<1\leq g_{\max}$. Their combined control at step $k$ is
\begin{equation}
\gamma_e^{(k)}
=
m_e^{(k)}d_e^{(k)}g_e^{(k)}.
\end{equation}

Let $\mathcal{I}_j$ denote the incoming edges of $V_j$, and let $x_e$ be the
state of the parent endpoint of $e$. The controls parameterize the CPT as
\begin{equation}
\label{eq:controlled_cpt}
\theta_{j,v\mid\mathbf{pa}_j}^{(k)}
=
\operatorname{softmax}_{v}
\left[
b_j(v)+
\sum_{e\in\mathcal{I}_j}
\gamma_e^{(k)}\psi_e(v,x_e)
\right],
\end{equation}
where $b_j$ is a base potential and $\psi_e$ is the contribution of edge $e$.
Collecting the CPTs gives
$B^{(k)}=(\mathcal{G}_{\mathrm{aug}},\Theta^{(k)})$.

For an accepted LLM edge, define
\begin{equation}
\bar g_e
=
g_{\min}+(1-g_{\min})c_e^{\mathrm{LLM}}.
\end{equation}
The initial controls are
\begin{equation}
(m_e^{(0)},g_e^{(0)},d_e^{(0)})
=
\begin{cases}
(1,1,1), & e\in\mathcal{E}_s,\\
(1,\bar g_e,1), & e\in\mathcal{E}^{+}.
\end{cases}
\end{equation}

We initialize the base and edge potentials in Eq.~\eqref{eq:controlled_cpt} to obtain the initial controlled BN $B^{(0)}$ by approximating $B^{\mathrm{aug}}$ under the controlled parameterization. Each episode starts from $B^{(0)}$ and runs for $T$ steps, where the policy updates one edge control at each step and evaluates the resulting rollout; edge controls are reset between episodes, while the actor and critic parameters are retained.

\paragraph{State.}
Let $M=|\mathcal{E}_{\mathrm{aug}}|$ and index the edges as
$e_i=(V_{u_i}\rightarrow V_{v_i})$, $i=1,\ldots,M$. At step $k$, the state representation associated with edge $e_i$ is
\begin{equation}
\mathbf{x}_i^{(k)}
=
\left[
\mathbf{h}_{i,\mathrm{ctrl}}^{(k)}
\Vert
\mathbf{h}_{i,\mathrm{align}}^{(k)}
\Vert
\mathbf{h}_{i,\mathrm{struct}}
\Vert
\mathbf{h}_{i,\mathrm{hist}}^{(k)}
\right],
\end{equation}
where the four blocks encode the current controls, generated--target
discrepancies, structural information, and operation history, respectively.
Let
\begin{equation}
\mathbf{X}^{(k)}
=
\left[
\mathbf{x}_1^{(k)},\ldots,\mathbf{x}_M^{(k)}
\right]^{\top},
\quad
s^{(k)}
=
\left(
\mathbf{X}^{(k)},\mathbf{A}_{\mathcal{E}}
\right),
\end{equation}
where $\mathbf{A}_{\mathcal{E}}$ connects two edge representations when their
BN edges share a variable. A graph encoder aggregates information among
connected edges; the actor scores valid edge--operation pairs, and the critic
estimates the state value.

\paragraph{Action and transition.}
The policy samples
\begin{equation}
a^{(k)}
=
(e^{(k)},o^{(k)})
\sim
\pi_{\phi}(\cdot\mid s^{(k)}),
\end{equation}
where
\begin{equation}
\begin{aligned}
\mathcal{O}
=
\{
\mathrm{keep},
\mathrm{weaken},
\mathrm{strengthen},
\mathrm{deactivate},
\mathrm{flip}
\}.
\end{aligned}
\end{equation}
For $0<\eta_-<1<\eta_+$, define
\begin{equation}
\begin{aligned}
g_e^{\downarrow}=
\max\{g_{\min},\eta_-g_e\},\,
g_e^{\uparrow}
=
\min\{g_{\max},\eta_+g_e\}.
\end{aligned}
\end{equation}
For brevity, the quantities on the right-hand side below denote the controls
at step $k$. The selected operation applies
\begin{equation}
(m_e^{(k+1)},g_e^{(k+1)},d_e^{(k+1)})
=
\begin{cases}
(m_e,g_e,d_e),
& \mathrm{keep},\\
(m_e,g_e^{\downarrow},d_e),
& \mathrm{weaken},\\
(1,g_e^{\uparrow},d_e),
& \mathrm{strengthen},\\
(0,g_e,d_e),
& \mathrm{deactivate},\\
(m_e,g_e,-d_e),
& \mathrm{flip}.
\end{cases}
\end{equation}
Deactivation preserves magnitude and polarity so that
\textit{strengthen} can reactivate the edge. The \textit{flip} operation
reverses the polarity of the edge-specific log-potential but does not change
the graph direction $V_u\rightarrow V_v$. Operations that cannot alter the
current controls are masked, while \textit{keep} remains available. After each action, Eq.~\eqref{eq:controlled_cpt} yields $B^{(k+1)}$; we sample $N_{\mathrm{roll}}$ records and recompute the dynamic state features.


\paragraph{Reward.}
The global rollout loss combines distributional fidelity, downstream utility,
and feature--label preservation:
\begin{equation}
\mathcal{J}^{(k)}
=
\lambda_{\mathrm{d}}\widehat D_{\mathrm{dist}}^{(k)}
+
\lambda_{\mathrm{u}}\widehat\Delta_U^{(k)}
+
\lambda_{\mathrm{a}}\widehat\Delta_A^{(k)}.
\end{equation}
All terms compare $\widehat{\mathcal{D}}_t^{(k)}$ with $\mathcal{D}_t^{\mathrm{fs}}$. For an edge
$e=(V_u\rightarrow V_v)$, its local conditional discrepancy is
\begin{equation}
\rho_e^{(k)}
=
\frac{1}{2}
\sum_{x\in\mathrm{Val}(V_u)}
\widehat p_t(x)
\sum_{v\in\mathrm{Val}(V_v)}
\left|
\widehat p_k(v\mid x)
-
\widehat p_t(v\mid x)
\right|,
\end{equation}
where $\widehat p_k$ and $\widehat p_t$ are smoothed estimates from the current
rollout and few-shot target table. The step reward is
\begin{equation}
\begin{aligned}
r^{(k)}
={}&
w_{\mathrm{g}}
\left(
\mathcal{J}^{(k)}-\mathcal{J}^{(k+1)}
\right)
+
w_{\mathrm{l}}
\left(
\rho_{e^{(k)}}^{(k)}
-
\rho_{e^{(k)}}^{(k+1)}
\right),
\end{aligned}
\end{equation}
where the global term evaluates the complete generated table and the local
term attributes improvement to the selected edge.

\paragraph{PPO update and synthesis.}
PPO uses generalized advantage estimation and the clipped actor objective,
while the critic is trained on the corresponding returns. Entropy
regularization encourages exploration, and a target KL threshold limits policy
drift. Across all rollouts, we retain the control configuration
$(\mathbf{m}^{\star},\mathbf{g}^{\star},\mathbf{d}^{\star})$ with the lowest
global loss. It defines
\begin{equation}
\begin{aligned}
\mathcal{E}^{\star}
=
\left\{
e\in\mathcal{E}_{\mathrm{aug}}:
m_e^{\star}=1
\right\}&,
\mathcal{G}^{\star}
=
(\mathcal{V},\mathcal{E}^{\star}),\\
\Theta^{\star}
=
\Theta(
\mathbf{m}^{\star},
\mathbf{g}^{\star},
\mathbf{d}^{\star}).
\end{aligned}
\end{equation}
The resulting BN
$B^{\star}=(\mathcal{G}^{\star},\Theta^{\star})$
is used to independently sample the $N$ synthetic target records specified in
the problem formulation.

\section{Experiments}

\begin{table*}[t]
\centering
\small
\setlength{\tabcolsep}{3.2pt}
\renewcommand{\arraystretch}{1.02}

\begin{tabular}{@{}lrrrrrrr@{}}
\hline
\multicolumn{8}{c}{\textit{(a) Overall by scenario} ($\downarrow$)} \\
\hline
Method & \textsc{inc-prsd} & \textsc{inc-edu} & \textsc{cov-txca} & \textsc{cov-edu} & \textsc{mob-prca} & \textsc{mob-sec} & Avg. \\
\hline
Gaussian Copula & 0.3200 & 0.3364 & 0.0729 & 0.0858 & 0.1096 & 0.2584 & $0.1972 \pm 0.0082$ \\
MTabGen & 0.1483 & \textbf{0.1069} & 0.0608 & 0.0599 & 0.2619 & 0.2231 & $0.1435 \pm 0.0087$ \\
CTGAN & 0.2200 & 0.1578 & 0.0767 & 0.0862 & 0.1318 & 0.1573 & $0.1383 \pm 0.0069$ \\
TVAE & 0.1770 & 0.2750 & 0.1082 & 0.0809 & 0.1743 & \textbf{0.0882} & $0.1506 \pm 0.0073$ \\
CopulaGAN & 0.2149 & 0.1311 & 0.0682 & 0.0693 & 0.3094 & 0.1189 & $0.1520 \pm 0.0079$ \\
SPADA & 0.1673 & 0.1230 & 0.2131 & 0.2056 & 0.1143 & 0.2059 & $0.1715 \pm 0.0170$ \\
\textbf{LAB-Tab} & \textbf{0.1367} & 0.1184 & \textbf{0.0243} & \textbf{0.0430} & \textbf{0.0837} & 0.1437 & $\boldsymbol{0.0916 \pm 0.0058}$ \\
\hline
\end{tabular}

\vspace{3pt}

\begin{tabular}{@{}lrrrr@{}}
\hline
\multicolumn{5}{c}{\textit{(b) Macro-averaged component metrics over 6 scenarios}} \\
\hline
Method & JSD $\downarrow$ & WAPE $\downarrow$ & Utility Gap $\downarrow$ & Coef--Cos $\uparrow$ \\
\hline
Gaussian Copula & $0.0195 \pm 0.0003$ & $0.3061 \pm 0.0024$ & $0.3986 \pm 0.0120$ & $0.3783 \pm 0.0065$ \\
MTabGen & $0.0213 \pm 0.0006$ & $0.2753 \pm 0.0039$ & $0.0751 \pm 0.0045$ & $0.4740 \pm 0.0069$ \\
CTGAN & $0.0265 \pm 0.0007$ & $0.3097 \pm 0.0075$ & $0.1458 \pm 0.0074$ & $0.6816 \pm 0.0053$ \\
TVAE & $0.0492 \pm 0.0034$ & $0.3568 \pm 0.0097$ & $0.1556 \pm 0.0351$ & $\boldsymbol{0.6919 \pm 0.0147}$ \\
CopulaGAN & $0.0332 \pm 0.0017$ & $0.3259 \pm 0.0058$ & $0.1185 \pm 0.0076$ & $0.5620 \pm 0.0062$ \\
SPADA & $0.0167 \pm 0.0006$ & $0.2317 \pm 0.0040$ & $0.1381 \pm 0.0074$ & $0.2747 \pm 0.0154$ \\
\textbf{LAB-Tab} & $\boldsymbol{0.0016 \pm 0.0003}$ & $\boldsymbol{0.1600 \pm 0.0024}$ & $\boldsymbol{0.0728 \pm 0.0037}$ & $0.6464 \pm 0.0458$ \\
\hline
\end{tabular}
\caption{Results at the 10\% target-data budget. Panel (a) reports
scenario-level Overall scores, and Panel (b) reports macro-averaged component
metrics. Values are averaged over five seeds; macro results are mean $\pm$
standard deviation. Scenario-level standard deviations are provided in
Appendix F. Best values are bold.}
\label{tab:main-results}
\end{table*}

\subsection{Experimental Setup}

\paragraph{Data and scenarios.}
We construct six source--target shifts from three 2018 ACS prediction tasks in Folktables \citep{ding2021retiring}. The regression scenarios are \textsc{INC-PRSD} (ACSIncome, Puerto Rico $\rightarrow$ South Dakota) and \textsc{INC-EDU} (ACSIncome, non-bachelor $\rightarrow$ bachelor-or-above in California). The binary-classification scenarios are \textsc{COV-TXCA} (ACSPublicCoverage, Texas $\rightarrow$ California), \textsc{COV-EDU} (ACSPublicCoverage, non-bachelor $\rightarrow$ bachelor-or-above in California), \textsc{MOB-PRCA} (ACSMobility, Puerto Rico $\rightarrow$ California), and \textsc{MOB-SEC} (ACSMobility, private-sector $\rightarrow$ government workers in California). For each scenario, we reserve 20\% of the target records for testing. The remaining records form the target training set used for the main experiment and the sensitivity analysis as provided in Appendix~A.

\paragraph{Metrics.}
All metrics are computed against held-out real target data. \textbf{JSD} is the mean marginal Jensen--Shannon divergence, and define
\begin{equation}
\textbf{WAPE}=\frac{\sum_i|y_i-\hat{y}_i|}{\sum_i|y_i|},
\end{equation}
where $y_i$ is the held-out target label and $\hat{y}_i$ is the predicted income for \textsc{INC-PRSD} and \textsc{INC-EDU}, or the predicted positive-class probability for the four classification scenarios. Let $f_r$ and $f_s$ denote identical predictors trained on the real few-shot and synthetic target data, respectively. We define
\begin{equation}
\textbf{UtilityGap}
=|U(f_r;\mathcal{D}_t^{\mathrm{test}})
-U(f_s;\mathcal{D}_t^{\mathrm{test}})|,
\end{equation}
where $U$ is $R^2$ for regression and ROC-AUC for classification. Because the predictor architecture and test set are fixed, UtilityGap measures how closely synthetic training data reproduce the predictive utility of real
few-shot target data. After applying a shared one-hot encoding to categorical features and standardizing numerical features, we compute
\begin{equation}
\textbf{Coef--Cos} = \frac{\boldsymbol{\beta}_r^\top\boldsymbol{\beta}_s}{
\|\boldsymbol{\beta}_r\|_2\,\|\boldsymbol{\beta}_s\|_2},
\end{equation}
where $\boldsymbol{\beta}_r$ and $\boldsymbol{\beta}_s$ are the coefficient vectors of linear probes fitted to real and synthetic target data. JSD, WAPE, and UtilityGap are lower-is-better, whereas Coef--Cos is
higher-is-better. Let $J$, $W$, $G$, and $C$ denote the four metrics for one
scenario and seed. We define
\begin{equation}
\begin{aligned}
\textbf{Overall}
=
\frac{1}{4}\bigg(
\frac{J}{\ln 2}
+\frac{W}{1+W}
+\frac{G}{1+G}
+\frac{1-C}{2}
\bigg).
\end{aligned}
\end{equation}
Overall is bounded in $[0,1]$ and is lower-is-better.

\paragraph{Baselines and implementation.}
We compare CopulaGAN \citep{patki2016sdv}, CTGAN \citep{xu2019ctgan}, TVAE \citep{xu2019ctgan}, Gaussian Copula \citep{asghar2019dpgaussiancopula}, MTabGen \citep{villaizan2025mtabgen}, and SPADA \citep{yang2025spada}. All methods receive the same source data and few-shot target subset and are evaluated on the same held-out target split. The BN is learned by BIC hill climbing with $d_{\max}=4$, $\alpha=10$, and $\kappa=1$. PPO runs for 150 episodes on regression scenarios and 250 episodes on classification scenarios, with 20 steps per episode, a clip ratio of $0.2$, and a learning rate of $3\times10^{-4}$. Candidate edges are proposed using \textit{gpt-4o}. The 10\% main-table results are averaged over five random seeds, and uncertainty denotes the empirical standard deviation across seeds, more implementation details are given in Appendix~B and ~C.

\subsection{Main Results}

\paragraph{Comparison at the 10\% few-shot budget.}
Table~\ref{tab:main-results} shows that LAB-Tab achieves the lowest
macro Overall score, $0.0916\pm0.0058$, which is 33.8\% lower than CTGAN,
the strongest baseline at $0.1383\pm0.0069$. LAB-Tab ranks first in four of
the six scenarios, while MTabGen and TVAE obtain the lowest Overall scores on
\textsc{INC-EDU} and \textsc{MOB-SEC}, respectively. This advantage is
supported by the best macro JSD, WAPE, and UtilityGap, together with
competitive Coef--Cos. The baselines exhibit sharper trade-offs: SPADA provides
the strongest baseline distributional alignment but preserves less downstream
utility and feature--label association; MTabGen approaches LAB-Tab in
UtilityGap but remains weaker in distributional fidelity and task-specific
error; and TVAE and CTGAN obtain higher Coef--Cos but incur larger
discrepancies on the other components. LAB-Tab's leading Overall score
therefore reflects consistent performance across different objectives. 

\begin{figure*}[t]
\centering
\includegraphics[width=0.8\textwidth]{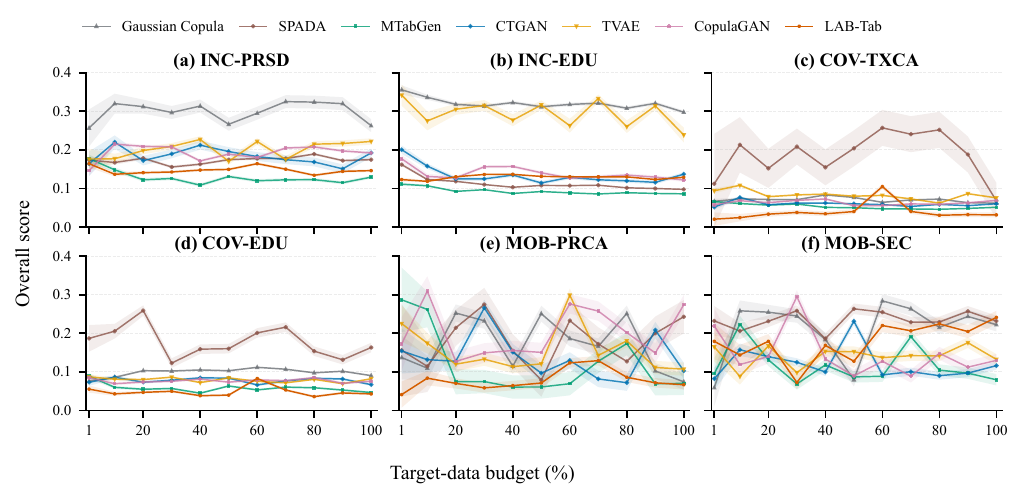}
\caption{Macro Overall as the target-data fraction increases from 1\% to 100\%. Each method is retrained at every fraction, and lower values indicate better joint performance. Shading denotes variation across five random seeds. The structural adaptation cases and policy-ablation training dynamics are available in Appendices~D and~E.}
\label{fig:data-fraction-curves}
\end{figure*}

\paragraph{Sensitivity to the target-data fraction.}
Figure~\ref{fig:data-fraction-curves} shows that LAB-Tab has the lowest macro Overall at seven of the eleven target-data fractions and the lowest average score across the sweep. Its advantage is concentrated in the low-data regime: at 1\%, LAB-Tab scores 0.1499 versus 0.3299 for CTGAN, a 54.6\% reduction. The gap narrows as target evidence grows, and MTabGen leads at 40\%, 60\%, 90\%, and 100\%; even at the full-data endpoint, however, LAB-Tab remains close (0.1751 versus 0.1650). This crossover is consistent with source-informed structural adaptation being most valuable when target observations are scarce, whereas MTabGen benefits more from direct target evidence. LAB-Tab's stable trajectory further shows that its few-shot advantage does not come with a sharp loss of competitiveness at larger budgets. 

\begin{table}[t]
\centering
\footnotesize
\setlength{\tabcolsep}{2.4pt}
\renewcommand{\arraystretch}{1.04}
\begin{tabular}{@{}lrrrrr@{}}
\toprule
Method & JSD $\downarrow$ & WAPE $\downarrow$ & UGap $\downarrow$ & Coef. $\uparrow$ & Overall $\downarrow$ \\
\midrule
Source BN & 0.0318 & 0.3137 & 0.1861 & 0.5280 & 0.1633 \\
Target BN & \textbf{0.0011} & 0.1815 & 0.1550 & 0.4137 & 0.1382 \\
Source BN + MAP & \textbf{0.0011} & 0.1641 & 0.0923 & 0.6082 & 0.1003 \\
\midrule
w/o LLM & 0.0015 & 0.1659 & 0.0929 & 0.4478 & 0.1211 \\
w/o PPO & 0.0013 & 0.1626 & 0.0766 & 0.5803 & 0.0997 \\
Greedy edits & 0.0015 & 0.1707 & 0.0995 & 0.5536 & 0.1087 \\
w/o local reward & 0.0016 & \textbf{0.1584} & 0.0865 & 0.4892 & 0.1137 \\
\midrule
\textbf{LAB-Tab} & 0.0016 & 0.1600 & \textbf{0.0728} & \textbf{0.6464} & \textbf{0.0916} \\
\bottomrule
\end{tabular}
\parbox{\columnwidth}{\scriptsize
UGap denotes Utility Gap and Coef.\ denotes Coef--Cos.}
\caption{Ablation results averaged over five random seeds and six drift scenarios.}
\label{tab:ablation-main}
\end{table}

\subsection{Ablation Study}

\begin{figure}[t]
    \centering
    \includegraphics[width=0.8\linewidth]{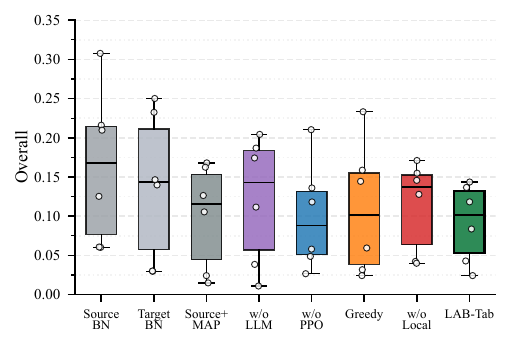}
    \caption{Scenario-wise Overall distributions for the ablations. Each
    circle is the seed-averaged score for one scenario; boxes show the
    interquartile range and horizontal lines show the median.}
    \label{fig:ablation-overall}
\end{figure}

\paragraph{Ablation settings.}
All variants use identical 10\% target splits, seeds, and evaluation protocol. Source BN reuses the source-fitted graph and parameters, Target BN is fitted entirely on few-shot target data, and Source BN + MAP retains the source graph while updating its conditional probabilities with target counts. The w/o LLM, w/o PPO, and w/o local reward variants remove proposed edges, policy control, and edge-level feedback, respectively; Greedy edits replaces PPO with locally selected one-step actions. 

\paragraph{Component contributions.}
Table~\ref{tab:ablation-main} reveals distinct roles for the components of LAB-Tab. Direct source reuse performs poorly under target shift, while the target-only and MAP variants show that fitting sparse target data or they can match target marginals yet fail to preserve predictive and feature--label relationships. Removing LLM augmentation leaves JSD nearly unchanged but substantially reduces Coef--Cos, indicating that the proposed edges mainly contribute target-specific conditional structure. PPO then coordinates these dependencies: disabling it or replacing it with greedy edits produces less stable performance across scenarios, as reflected by the higher upper tails in Figure~\ref{fig:ablation-overall}. Removing the local reward slightly improves WAPE because the policy can focus more directly on the global prediction objective, but this also weakens conditional and feature--label alignment. Similarly, the full model's slightly higher JSD reflects a small marginal-fidelity trade-off for better dependency preservation and downstream utility. The results show that LLM augmentation expands the target dependency space, while PPO coordination prevents optimization of aggregate errors from distorting conditional relationships.

\section{Conclusion}
Few-shot target-domain synthesis must reuse transferable source dependencies without retaining relationships that shift across domains. \textbf{LAB-Tab} addresses this challenge by adapting an explicit BN, it learns a source structure, introduces target-specific candidate edges under graph-validity constraints, recalibrates the augmented BN with few-shot target counts, and applies PPO to control edge activation, magnitude, and polarity through global and local feedback. Across six ACS source--target shifts at the 10\% target-data budget, LAB-Tab achieves the lowest Overall score in four scenarios and reduces macro Overall by 33.8\% relative to the strongest baseline, while obtaining the best macro JSD, WAPE, and Utility Gap and remaining competitive in feature--label preservation. It also leads at seven of eleven target-data fractions, with its largest gains at 1\% and 10\%. Ablations show that LLM augmentation, PPO-based edge control, and local dependency feedback provide complementary benefits beyond source reuse and MAP recalibration. Overall, LAB-Tab provides an effective and inspectable approach to few-shot tabular generation by combining explicit structural hypotheses with feedback-driven source-to-target adaptation.

\newpage
\bibliography{references}
\input{appendix}
\end{document}

%% file: appendix.tex
\clearpage
\makeatletter
\setlength{\@dblfptop}{0pt}
\setlength{\@dblfpsep}{8pt plus 2pt minus 2pt}
\setlength{\@dblfpbot}{0pt plus 1fil}
\makeatother
\appendix

\section{Dataset and Scenario Details}
\label{app:data}

\subsection{ACS Task Schemas}

All experiments use person-level records from the 2018 ACS distributed through
Folktables. Income retains working-age individuals with positive personal
income and weekly working hours. Coverage focuses on individuals younger than
65 with annual income no greater than \$30,000, and Mobility focuses on adults
aged 19--34. The targets are personal income, public health insurance
coverage, and residential mobility, respectively.

\begin{table*}[!t]
\caption{Modeled variables for the three ACS tasks. Cat.\ denotes categorical,
Q-num.\ quantile-binned numeric, Q-code quantile-binned code, and Ord.\
source-defined ordinal bins. Bold variables are labels; all remaining
variables are features. State counts can be smaller when quantile boundaries
coincide.}
\label{tab:app-schema}
\centering
\scriptsize
\setlength{\tabcolsep}{2.2pt}
\renewcommand{\arraystretch}{0.92}
\begin{tabular*}{\textwidth}{@{\extracolsep{\fill}}llcllcllc@{}}
\toprule
\multicolumn{3}{c}{Income}
& \multicolumn{3}{c}{Coverage}
& \multicolumn{3}{c}{Mobility} \\
\cmidrule(lr){1-3}\cmidrule(lr){4-6}\cmidrule(lr){7-9}
Variable & Type & States
& Variable & Type & States
& Variable & Type & States \\
\midrule
Age & Q-num. & 8
& Age & Q-num. & 8
& Age & Q-num. & 8 \\
Class of worker & Cat. & 8
& Education & Cat. & 24
& Education & Cat. & 24 \\
Education & Cat. & 24
& Marital status & Cat. & 5
& Marital status & Cat. & 5 \\
Marital status & Cat. & 5
& Sex & Cat. & 2
& Sex & Cat. & 2 \\
Occupation & Q-code & 8
& Disability & Cat. & 2
& Disability & Cat. & 2 \\
Household relationship & Cat. & 17--18
& Citizenship & Cat. & 5
& Citizenship & Cat. & 5 \\
Weekly working hours & Q-num. & 5
& Prior-year mobility & Cat. & 3
& Class of worker & Cat. & 10 \\
Sex & Cat. & 2
& Personal income & Q-num. & 7
& Weekly working hours & Q-num. & 6 \\
Race & Cat. & 8--9
& Employment status & Cat. & 7
& Personal income & Q-num. & 6--8 \\
\textbf{Personal income} & Ord. & 10
& Race & Cat. & 9
& Race & Cat. & 9 \\
& &
& \textbf{Public coverage} & Binary & 2
& \textbf{Residential mobility} & Binary & 2 \\
\bottomrule
\end{tabular*}
\end{table*}

\subsection{Source--Target Scenario Construction}

For each task, we construct one spatial shift and one population-group shift.
The spatial scenarios are Puerto Rico to South Dakota for Income, Texas to
California for Coverage, and Puerto Rico to California for Mobility. The
population-group scenarios are constructed within California: the education
shifts compare non-bachelor and bachelor-or-above groups, and the sector shift
compares private-sector and government workers.
\begin{table}[!t]
\caption{Dataset sizes for the six source--target scenarios.
$n_{\mathrm{adapt}}$ is the target set available for adaptation, and $d_x$
excludes the label.}
\label{tab:app-scenarios}
\centering
\footnotesize
\setlength{\tabcolsep}{4.0pt}
\renewcommand{\arraystretch}{1.02}
\begin{tabular}{@{}lrrrr@{}}
\toprule
Scenario & $n_s$ & $n_{\mathrm{adapt}}$ & $n_{\mathrm{test}}$ & $d_x$ \\
\midrule
\textsc{inc-prsd} & 9,071   & 1,595 & 980    & 9  \\
\textsc{inc-edu}  & 120,016 & 6,083 & 15,128 & 9  \\
\textsc{cov-txca} & 98,928  & 7,841 & 27,711 & 10 \\
\textsc{cov-edu}  & 115,947 & 3,416 & 4,521  & 9  \\
\textsc{mob-prca} & 4,730   & 1,707 & 16,066 & 10 \\
\textsc{mob-sec}  & 56,723  & 2,064 & 1,678  & 9  \\
\bottomrule
\end{tabular}
\end{table}

\subsection{Data Splits and Target-Data Budgets}

For each reporting seed, a stratified split reserves 20\% of the target
domain for held-out testing. A scenario-specific adaptation set is then
sampled from the remaining records, with the sizes reported in
Table~\ref{tab:app-scenarios}. All methods use the same test and adaptation
records within a scenario and seed.

A seeded permutation produces nested adaptation subsets containing
$\{1,10,20,\ldots,100\}\%$ of the adaptation set. The default experiment and
the 10\% sensitivity point use the same subset and run configuration.

\subsection{Preprocessing and Discretization}

The BN operates on finite state spaces. Each non-label numerical variable is
discretized into at most eight source-domain quantile bins, and the same
boundaries are applied to the target data. Repeated boundaries are merged, and
missing feature values form an explicit category.

For Income, the label is transformed by $\log(1+\mathrm{income})$ and
discretized into ten source-defined ordinal states for BN modeling. Each state is decoded
through its source-bin median when regression metrics are computed in dollars.
Coverage and Mobility use binary labels.

\section{Evaluation Metrics and Aggregation}
\label{app:metrics}

Let $\mathcal{D}_t^{\mathrm{test}}$ denote the held-out real target table and $\widehat{\mathcal{D}}_t$ a synthetic target table. All headline metrics use the same processed variables and are evaluated against $\mathcal{D}_t^{\mathrm{test}}$.

\subsection{Distributional Fidelity and WAPE}

\paragraph{Mean marginal JSD.} For a variable $V_j$ with state space $\mathcal{A}_j$, let $p_j(a)$ and $\widehat p_j(a)$ denote the empirical probabilities of state $a$ in $\mathcal{D}_t^{\mathrm{test}}$ and $\widehat{\mathcal{D}}_t$, respectively, and let $m_j=(p_j+\widehat p_j)/2$. The marginal Jensen--Shannon divergence is
\begin{equation}
\mathrm{JSD}_j=\frac{1}{2}\mathrm{KL}(p_j\|m_j)+\frac{1}{2}\mathrm{KL}(\widehat p_j\|m_j),
\end{equation}
where natural logarithms are used and zero-probability terms contribute zero. We average this quantity over all modeled variables, including the label:
\begin{equation}
\mathrm{JSD}=\frac{1}{|\mathcal{V}|}\sum_{V_j\in\mathcal{V}}\mathrm{JSD}_j.
\end{equation}
All variables are evaluated in their discrete BN representations; the Income label is not decoded for JSD.

\paragraph{WAPE.} For each scenario, a task predictor is trained on the synthetic target table and evaluated on the held-out real target records. We compute
\begin{equation}
\mathrm{WAPE}
=
\frac{\sum_{i=1}^{n_{\mathrm{test}}}|y_i-\widehat y_i|}
{\sum_{i=1}^{n_{\mathrm{test}}}|y_i|},
\end{equation}
where $y_i$ is the held-out target label. For \textsc{inc-prsd} and \textsc{inc-edu}, $\widehat y_i$ is the predicted income after decoding the ordinal income states to their source-bin medians. For \textsc{cov-txca}, \textsc{cov-edu}, \textsc{mob-prca}, and \textsc{mob-sec}, $\widehat y_i$ is the predicted positive-class probability. The same WAPE definition is used in all six scenarios.

\subsection{Downstream Utility Gap}

We evaluate downstream utility using Train on Synthetic, Test on Real (TSTR) and Train on Real, Test on Real (TRTR) under a matched target-test protocol. The TSTR predictor is trained on $\widehat{\mathcal{D}}_t$, while the TRTR reference predictor is trained on the same few-shot target subset available to the generator. Both predictors use the same model family and are evaluated on the same fixed $\mathcal{D}_t^{\mathrm{test}}$. With $U_{\mathrm{TSTR}}$ and $U_{\mathrm{TRTR}}$ denoting the corresponding test utilities, we report
\begin{equation}
\mathrm{UtilityGap}=|U_{\mathrm{TRTR}}-U_{\mathrm{TSTR}}|.
\end{equation}
For Income, $U$ is the coefficient of determination $R^2$ after decoding
the ordinal income states. For Coverage and Mobility, $U$ is ROC-AUC computed
from positive-class probabilities. A smaller UtilityGap indicates that
training on synthetic data more closely reproduces the predictive utility of
the real few-shot target reference.

For each task, discrete feature states are ordinal-encoded. To keep feature
codes aligned across real and synthetic evaluation tables, the encoder
vocabulary is formed from the observed training and held-out feature states
without using held-out labels. Any state outside this vocabulary is mapped to
$-1$. Hyperparameters are fixed across methods, all predictor fits use fixed
seed 0, and no predictor is tuned separately for a generator.
Table~\ref{tab:app-predictors} summarizes the configurations.

\begin{table*}[!t]
\caption{Downstream predictors used for TSTR and TRTR evaluation.}
\label{tab:app-predictors}
\centering
\footnotesize
\setlength{\tabcolsep}{4pt}
\renewcommand{\arraystretch}{1.02}
\begin{tabular}{@{}lp{0.43\textwidth}p{0.38\textwidth}@{}}
\toprule
Task & Predictor and input & Configuration \\
\midrule
Income & Histogram gradient-boosting regressor; ordinal feature states; income states decoded to source-bin medians & 150 iterations; fixed seed 0; utility: $R^2$ \\
Coverage & Histogram gradient-boosting classifier; ordinal feature states; unseen states mapped to $-1$ & 200 iterations; depth 6; learning rate $0.1$; fixed seed 0; utility: ROC-AUC \\
Mobility & Histogram gradient-boosting classifier; ordinal feature states; unseen states mapped to $-1$ & 200 iterations; depth 6; learning rate $0.1$; fixed seed 0; utility: ROC-AUC \\
\bottomrule
\end{tabular}
\end{table*}

\subsection{Feature--Label Coefficient Similarity}

Coef--Cos evaluates whether a synthetic table preserves feature--label associations captured by matched linear probes. We apply a shared one-hot encoding to categorical features and standardize numerical features using a common preprocessing map. Two probes with the same specification are then fitted separately on the real and synthetic target tables, yielding coefficient vectors $\boldsymbol{\beta}_{\mathrm{real}}$ and $\boldsymbol{\beta}_{\mathrm{syn}}$. We report
\begin{equation}
\mathrm{Coef\text{-}Cos}
=
\frac{\boldsymbol{\beta}_{\mathrm{real}}^{\top}\boldsymbol{\beta}_{\mathrm{syn}}}
{\|\boldsymbol{\beta}_{\mathrm{real}}\|_2\,\|\boldsymbol{\beta}_{\mathrm{syn}}\|_2}.
\end{equation}
Higher values indicate closer agreement in coefficient direction and
stronger preservation of feature--label associations.

\subsection{Overall Score}

Overall combines JSD, WAPE, Utility Gap, and Coef--Cos on a fixed
loss-oriented scale. For scenario $s$, method $m$, and seed $r$, let
$J_{s,m}^{(r)}$, $W_{s,m}^{(r)}$, $G_{s,m}^{(r)}$, and $C_{s,m}^{(r)}$
denote the four component metrics. We define
\begin{equation}
\begin{aligned}
O_{s,m}^{(r)}
={}&
\frac{1}{4}\bigg(
\frac{J_{s,m}^{(r)}}{\ln 2}
+
\frac{W_{s,m}^{(r)}}{1+W_{s,m}^{(r)}}
\\
&\qquad
+
\frac{G_{s,m}^{(r)}}{1+G_{s,m}^{(r)}}
+
\frac{1-C_{s,m}^{(r)}}{2}
\bigg).
\end{aligned}
\end{equation}
Each term is bounded in $[0,1]$, and lower values indicate better
performance. The JSD term uses its theoretical upper bound under natural
logarithms; the WAPE and Utility Gap transformations preserve their ordering
while bounding potentially unbounded values; and the Coef--Cos term converts
similarity into a loss.

For each seed, the six-scenario macro score is
\begin{equation}
O_m^{(r)}
=
\frac{1}{6}
\sum_{s=1}^{6}
O_{s,m}^{(r)}.
\end{equation}
Because the transformation is fixed, Overall is independent of the methods
included in a table and is directly comparable across the headline,
target-budget, and ablation experiments.

\subsection{Aggregation and Uncertainty}

All headline experiments are repeated with $R=5$ independent reporting seeds. For a metric $Q$, the scenario-level mean and sample standard deviation are
\begin{equation}
\overline Q_{s,m}=\frac{1}{R}\sum_{r=1}^{R}Q_{s,m}^{(r)},
\qquad R=5,
\end{equation}
\begin{equation}
S_{s,m}=\sqrt{\frac{1}{R-1}\sum_{r=1}^{R}
\left(Q_{s,m}^{(r)}-\overline Q_{s,m}\right)^2}.
\end{equation}
For Overall, the fixed transformation is applied first within each seed and
scenario. The six-scenario average is then computed within each seed, followed
by the mean and empirical standard deviation across five seeds. Scenario
entries without uncertainty are five-seed means. The 10\% headline table,
10\% sensitivity point, and 10\% ablations follow this same order.
Appendix~\ref{app:ppo-dynamics} presents representative single-seed training
traces, while cross-seed uncertainty comes from the five reporting seeds.

\section{Implementation and Reproducibility Details}
\label{app:implementation}

\subsection{Bayesian-Network Learning and Calibration}

The source Bayesian network is learned independently for each scenario. Structure learning starts from an empty directed acyclic graph and applies greedy hill climbing with the Bayesian information criterion. To bound the cost of structure search, at most 10,000 source records are selected; parameter estimation still uses the complete source training table. The search evaluates single-edge addition, deletion, and reversal operations, rejects operations that introduce a cycle or violate the maximum in-degree, and terminates after 200 iterations or when no admissible operation improves the score. No random restart is used. A fixed column order and deterministic operator enumeration are retained throughout a run.

After structure learning, every conditional probability table (CPT) is estimated on the complete source table with a symmetric BDeu prior of equivalent sample size $\alpha=10$. Following LLM augmentation, the source CPTs are re-estimated on the augmented parent sets and calibrated with the few-shot target counts using $\kappa=1$. Unobserved parent configurations retain the source conditional distribution; observed configurations combine the source prior with target counts. Table~\ref{tab:app-bn-settings} summarizes the settings used in all six scenarios.

\begin{table}[!t]
\caption{Bayesian-network and sampling settings.}
\label{tab:app-bn-settings}
\centering
{
\small
\setlength{\tabcolsep}{4pt}
\begin{tabular}{@{}lp{0.55\columnwidth}@{}}
\toprule
Component & Setting \\
\midrule
Discretization & At most 8 source-quantile states per numerical variable \\
Structure score & BIC greedy hill climbing from an empty DAG \\
Structure rows & $\min(|\mathcal{D}_s|,10{,}000)$; fixed seed 0 \\
Search limit & 200 operations; no random restart \\
Maximum in-degree & $d_{\max}=4$ \\
Source CPT prior & BDeu, $\alpha=10$ \\
Target MAP prior & Source-centered Dirichlet, $\kappa=1$ \\
PPO rollout size & 800 rows for Income; 1,500 otherwise \\
\bottomrule
\end{tabular}
}

\end{table}

For each action comparison, the pre-action and post-action samplers use
common random numbers. This holds the underlying draws fixed while changing the
selected edge control and reduces Monte Carlo noise in the reward difference.
The final synthetic table is generated with a separate fixed seed after
selecting the best edge-control configuration.

\subsection{Edge-Control Parameterization}
\label{app:edge-control}

The base potential $b_j$ and edge potentials $\psi_e$ are initialized from
$\Theta_{\mathrm{aug}}$ to obtain the controlled BN $B^{(0)}$ used at the
start of each episode. These potentials remain fixed during PPO, while the
policy updates the activation $m_e$, magnitude $g_e$, and polarity $d_e$ of
every edge in $\mathcal{E}_{\mathrm{aug}}$. Thus, each action changes the
contribution of a validated dependency without altering its endpoints or
parent--child orientation.

During optimization, $\mathcal{G}_{\mathrm{aug}}$ defines the editable edge
set. Deactivation sets an edge contribution to zero and allows later
reactivation through \textit{strengthen}, while \textit{flip} changes its
polarity. The retained activation variables define
$\mathcal{G}^{\star}$, and the retained controls determine
$\Theta^{\star}$ for final sampling.

\subsection{PPO Architecture, Reward, and Selection}

Each augmented BN edge is represented by four feature blocks: current
controls, target-alignment statistics, local graph context, and adjustment
history. The alignment block contains edge-conditional and endpoint-level
generated--target discrepancies. Two edge representations are adjacent when their BN edges share an endpoint. A
two-layer GraphSAGE encoder maps the normalized coordinates to 64-dimensional
edge embeddings. The actor applies a 128-unit multilayer perceptron to each
edge embedding; the critic applies a separate 128-unit multilayer perceptron
to mean- and max-pooled graph embeddings.

At each step, the actor samples one valid edge--operation pair. Keep, weaken, strengthen, and deactivate control the selected contribution, and strengthen can reactivate a deactivated edge. The \textit{flip} action changes its polarity. All actions preserve the validated endpoints and parent--child orientation. Ineffective operations are masked, and keep remains valid for every edge.

Let $\widehat A_k$ and $R_k$ denote the generalized-advantage estimate and return for transition $k$. The actor and critic minimize
\begin{equation}
\begin{aligned}
\mathcal{L}_{\mathrm{PPO}}={}&-\mathcal{L}_{\mathrm{clip}}+c_V\mathbb{E}_k\!\left[(V(s_k)-R_k)^2\right]\\
&-c_H\mathbb{E}_k\!\left[H\!\left(\pi(\cdot\mid s_k)\right)\right],
\end{aligned}
\end{equation}
where $c_V$ is the value-loss coefficient and $c_H$ the
episode-dependent entropy coefficient. Advantages are standardized within an
episode. State coordinates are normalized online with Welford statistics and
clipped to $[-10,10]$; rewards are divided by their running standard deviation
and clipped to the same interval. Gradient norms are clipped at 0.5.

For each run, the state with the smallest global alignment loss over all
episodes and steps is retained. The held-out target set is excluded from both
policy optimization and configuration selection. Across the five reporting
seeds, the target split and few-shot subset are sampled independently, so the
reported variation reflects different target draws as well as model
randomness.

\begin{table}[!t]
\caption{PPO settings for the regression and classification scenarios.}
\label{tab:app-ppo-settings}
\centering
{
\small
\setlength{\tabcolsep}{3.5pt}
\begin{tabular}{@{}lcc@{}}
\toprule
Parameter & Income & Coverage/Mobility \\
\midrule
Episodes & 150 & 250 \\
Steps per episode & 20 & 20 \\
Rollout rows & 800 & 1,500 \\
Optimizer & Adam & Adam \\
Learning rate & $3\times10^{-4}$ & $3\times10^{-4}$ \\
PPO epochs per episode & 4 & 4 \\
Clip ratio $\epsilon$ & 0.20 & 0.20 \\
Discount factor $\gamma$ & 0.99 & 0.99 \\
GAE parameter $\lambda$ & 0.95 & 0.95 \\
Value coefficient $c_V$ & 0.50 & 0.50 \\
Entropy coefficient $c_H$ & $0.020\rightarrow0.003$ & $0.020\rightarrow0.003$ \\
Target KL threshold & 0.03 & 0.03 \\
Weaken factor $\eta_-$ & 0.85 & 0.85 \\
Strengthen factor $\eta_+$ & 1.20 & 1.20 \\
Global reward weight & 0.20 & 0.20 \\
Local reward weight & 1.00 & 1.00 \\
\bottomrule
\end{tabular}
}

\end{table}

\subsection{Baseline Configurations}

Every headline baseline receives the same source table and few-shot target
subset as LAB-Tab. The two tables are concatenated without a domain indicator,
target oversampling, or scenario-specific reweighting. The Source BN, Target
BN, and Source BN~+~MAP variants in
Table~\ref{tab:app-ablation-summary} provide complementary BN references.
All methods use the same source-fitted discretization and are evaluated on the
same target test set within each seed.

\begin{table*}[!t]
\caption{Headline baseline configurations. All methods use the same source table and few-shot target subset.}
\label{tab:app-baseline-settings}
\centering
\footnotesize
\setlength{\tabcolsep}{3.0pt}
\renewcommand{\arraystretch}{1.00}
\begin{tabular}{@{}lp{0.23\textwidth}p{0.58\textwidth}@{}}
\toprule
Method & Training data & Configuration \\
\midrule
Gaussian Copula & Source $+$ few-shot target & SDV Gaussian-copula synthesizer with all processed columns treated as categorical  \\
SPADA & Source $+$ few-shot target & Categorical specialization of dependency-aware sampling; LLM dependency DAG; temperature 1.0  \\
MTabGen & Source $+$ few-shot target & 80 epochs; batch 1,024; 3 transformer layers; 4 heads; width 96; learning rate $10^{-3}$; mask probability 0.35; 24 denoising steps  \\
CTGAN & Source $+$ few-shot target & SDV implementation; 100 epochs; package-default architecture and batch size; CPU training  \\
TVAE & Source $+$ few-shot target & SDV implementation; 100 epochs; package-default architecture and batch size; CPU training  \\
CopulaGAN & Source $+$ few-shot target & SDV implementation; 300 epochs; quantile-bin variables encoded ordinally during fitting and decoded after sampling  \\
\bottomrule
\end{tabular}
\end{table*}

MTabGen treats each discretized column as a categorical token with a column-specific vocabulary and optimizes masked-cell reconstruction using AdamW with weight decay $10^{-4}$. Sampling starts from empirical marginal draws and iteratively denoises the table at temperature 0.9. CopulaGAN maps quantile labels to ordinal numbers only for its internal copula transformation and maps generated values back to the nearest valid quantile state. SPADA constructs its LLM dependency graph separately for each scenario,
target-data budget, and reporting seed, followed by graph validation before
sampling.

\subsection{Computational Environment}

Experiments were executed on a Mac Studio with an Apple M2 Ultra processor (24 CPU cores) and 192~GB of unified memory. The archived environment uses Python 3.10.11, NumPy 2.0.2, pandas 2.3.3, SciPy 1.15.3, scikit-learn 1.7.2, pgmpy 1.1.2, Folktables 0.0.12, SDV 1.37.1, NetworkX 3.4.2, PyTorch 2.12.0, and PyTorch Geometric 2.8.0. LAB-Tab and the SDV baselines run on CPU in this environment. The scenario--seed LAB-Tab jobs were launched independently; measured wall-clock times from the training-event and result timestamps range from approximately 5.8 to 16.8 minutes per scenario.

Python, NumPy, and PyTorch are initialized separately for each of the five reporting seeds. The target permutation, model initialization, policy sampling, and synthetic sampling are derived deterministically from the corresponding seed.

LLM augmentation uses \textit{gpt-4o} with temperature 0.2 and a maximum output length of 4,096 tokens. A proposal is generated for each scenario, target-data budget, and reporting seed; validated proposals are reused across ablations with the same split and budget.

\section{LLM Edge Proposals and Structural Adaptation}
\label{app:llm}

\subsection{Prompt Construction}

The LLM receives aggregate context containing semantic variable names, state
domains, the source BN, source and target marginal summaries,
label-conditional summaries, and selected pairwise conditional contrasts.
The context is constructed from the source table and few-shot target subset.

The prompt requests a short ranked list of missing directed edges. Each
proposal contains a parent, child, confidence score, and concise rationale.
The LLM stage proposes structure; CPT estimation and row generation remain in
the BN pipeline.

\subsection{Proposal Validation}

Proposals are sorted by confidence and processed sequentially against the
updated graph. Proposals are validated to ensure valid variables, graph consistency, and DAG
constraints before being added to the source graph. Accepted proposals retain their confidence and rationale for initialization
and analysis. Malformed responses are retried, and the source BN is used when
no valid addition is returned.

\subsection{Proposal Statistics and Structural Cases}

Table~\ref{tab:app-llm-stats} reports the accepted additions shown in
Figure~\ref{fig:app-bn-edits} and representative PPO outcomes at the 10\%
target-data budget. For an accepted edge $e$, the LLM confidence initializes
its magnitude as
$g_e^{(0)}=g_{\min}+(1-g_{\min})c_e^{\mathrm{LLM}}$.

\begin{table*}[!t]
\centering
\caption{Structural adaptation statistics at the 10\% target-data budget.
Source edges denote dependencies in the source BN, LLM additions denote
new edges proposed by the LLM, and active additions denote proposed edges
retained after PPO calibration.}
\label{tab:app-llm-stats}
\small
\setlength{\tabcolsep}{3.0pt}
\renewcommand{\arraystretch}{1.05}

\begin{tabular}{@{}lccc p{0.55\textwidth}@{}}
\toprule
Scenario
& Source edges
& LLM additions
& Active additions
& LLM-proposed additions \\
\midrule

\textsc{inc-prsd}
& 8
& 3
& 3
& Age$\rightarrow$Income;
  Education$\rightarrow$Income;
  Marital status$\rightarrow$Income
\\

\textsc{inc-edu}
& 8
& 3
& 3
& Class of worker$\rightarrow$Income;
  Education$\rightarrow$Income;
  Occupation$\rightarrow$Income
\\

\textsc{cov-txca}
& 11
& 3
& 3
& Citizenship$\rightarrow$Coverage;
  Coverage$\rightarrow$Employment;
  Income$\rightarrow$Sex
\\

\textsc{cov-edu}
& 12
& 3
& 2
& Coverage$\rightarrow$Age;
  Coverage$\rightarrow$Sex;
  Citizenship$\rightarrow$Coverage
\\

\textsc{mob-prca}
& 8
& 5
& 4
& Age$\rightarrow$Mobility;
  Education$\rightarrow$Mobility;
  Marital status$\rightarrow$Mobility;
  Race$\rightarrow$Mobility;
  Citizenship$\rightarrow$Mobility
\\

\textsc{mob-sec}
& 9
& 3
& 2
& Race$\rightarrow$Mobility;
  Mobility$\rightarrow$Disability;
  Sex$\rightarrow$Mobility
\\

\bottomrule
\end{tabular}

\vspace{5pt}
\textbf{(b) Representative structural cases}\\[2pt]
\begin{tabular}{@{}llp{0.50\textwidth}lc@{}}
\toprule
Scenario & Accepted edge & Rationale & PPO outcome & Multiplier \\
\midrule
\textsc{inc-prsd}
& Age$\rightarrow$Income
& Age adds direct target-domain information for the income distribution.
& Strengthened & 1.20 \\
\textsc{inc-edu}
& Class of worker$\rightarrow$Income
& Class of worker captures income differences within the target education group.
& Weakened & 0.85 \\
\textsc{cov-txca}
& Coverage$\rightarrow$Employment
& Coverage contributes to target-specific employment patterns in the target domain.
& Flipped & $-1.00$ \\
\textsc{cov-edu}
& Citizenship$\rightarrow$Coverage
& Citizenship provides a plausible coverage-related dependency in the target group.
& Deactivated & 0.00 \\
\textsc{mob-prca}
& Marital status$\rightarrow$Mobility
& Marital status contributes to residential mobility under the geographic shift.
& Strengthened & 1.44 \\
\textsc{mob-sec}
& Race$\rightarrow$Mobility
& Race captures a target-specific mobility association under the sector shift.
& Weakened & 0.85 \\
\bottomrule
\end{tabular}
\end{table*}

LLM-proposed additions receive different controls across scenarios. PPO
strengthens, weakens, flips, or deactivates their contributions according to
the generated--target feedback.

\subsection{Scenario-Level BN Adaptation Patterns}
\label{app:bn-edits}

\begin{figure*}[!t]
\centering
\includegraphics[width=0.87\textwidth]{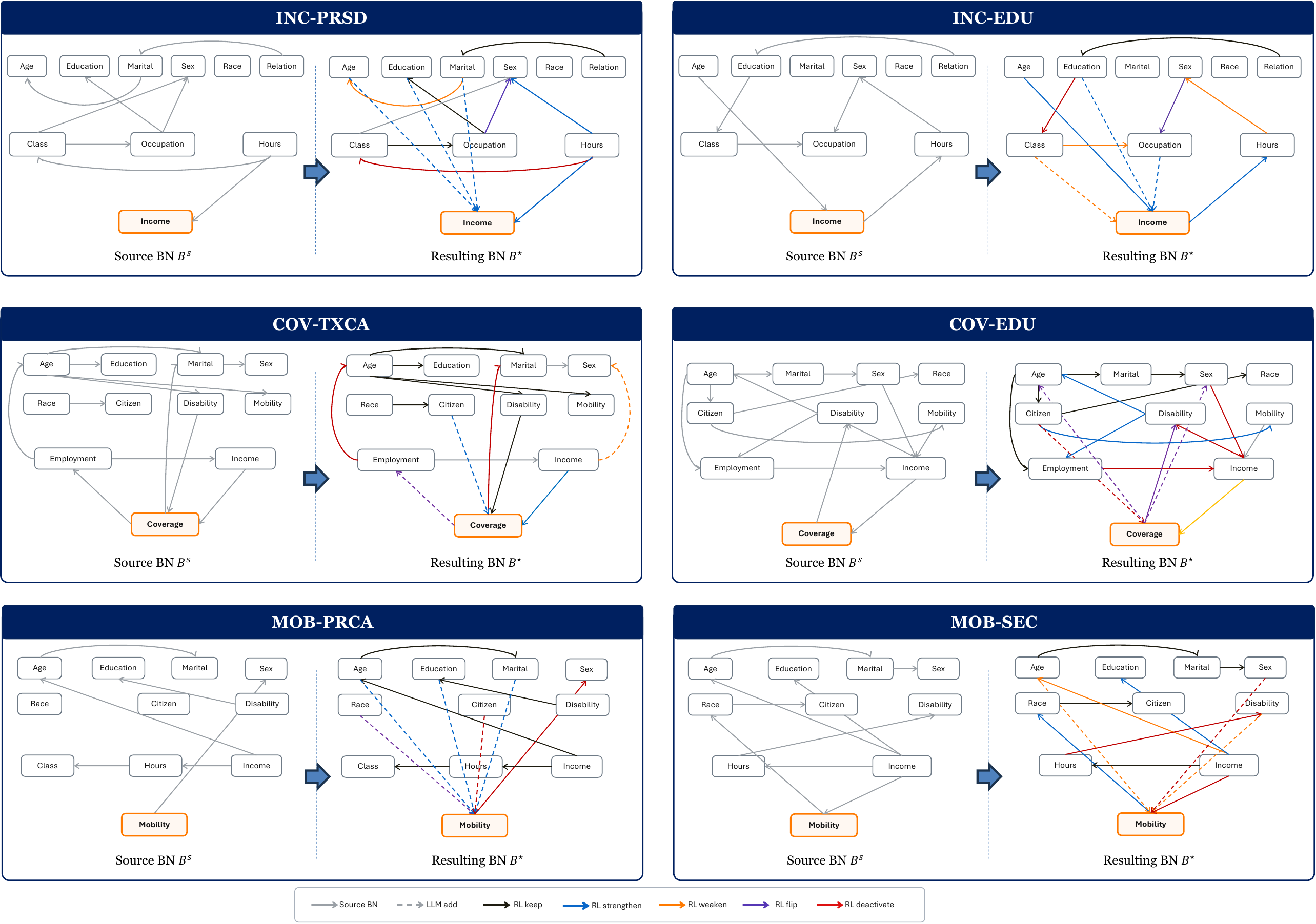}
\caption{Source and selected target-domain BN configurations for the six
shifts. Gray solid and dashed arrows denote source and LLM-added edges.
Black, blue, orange, purple, and red denote keep, strengthen, weaken, flip,
and deactivate outcomes. Deactivated edges are omitted from the final active
graph.}
\label{fig:app-bn-edits}
\end{figure*}

Figure~\ref{fig:app-bn-edits} shows that the source BN provides the structural
backbone in every scenario. LLM additions expand the editable edge set, and
PPO assigns the final controls to inherited and added dependencies. The
selected edits concentrate around label-related paths and differ across the
Income, Coverage, and Mobility shifts.

\section{PPO Optimization Dynamics}
\label{app:ppo-dynamics}

\subsection{Evolution of Edge-Control Actions}

For an episode $e$ containing $T=20$ edge-control steps, we summarize the policy behavior by the fraction of steps assigned to action $o$:
\begin{equation}
p_e(o)=\frac{1}{T}\sum_{k=1}^{T}\mathbf{1}\!\left[o^{(e,k)}=o\right].
\end{equation}
The five logged actions are keep, weaken, strengthen, deactivate, and flip. Deactivate sets the selected edge contribution to zero while leaving the augmented DAG unchanged. Flip changes the sign of the selected dependency contribution while preserving the parent--child orientation. These names follow the terminology used in the main paper.

\begin{figure*}[!t]
\centering
\includegraphics[width=0.87\textwidth]{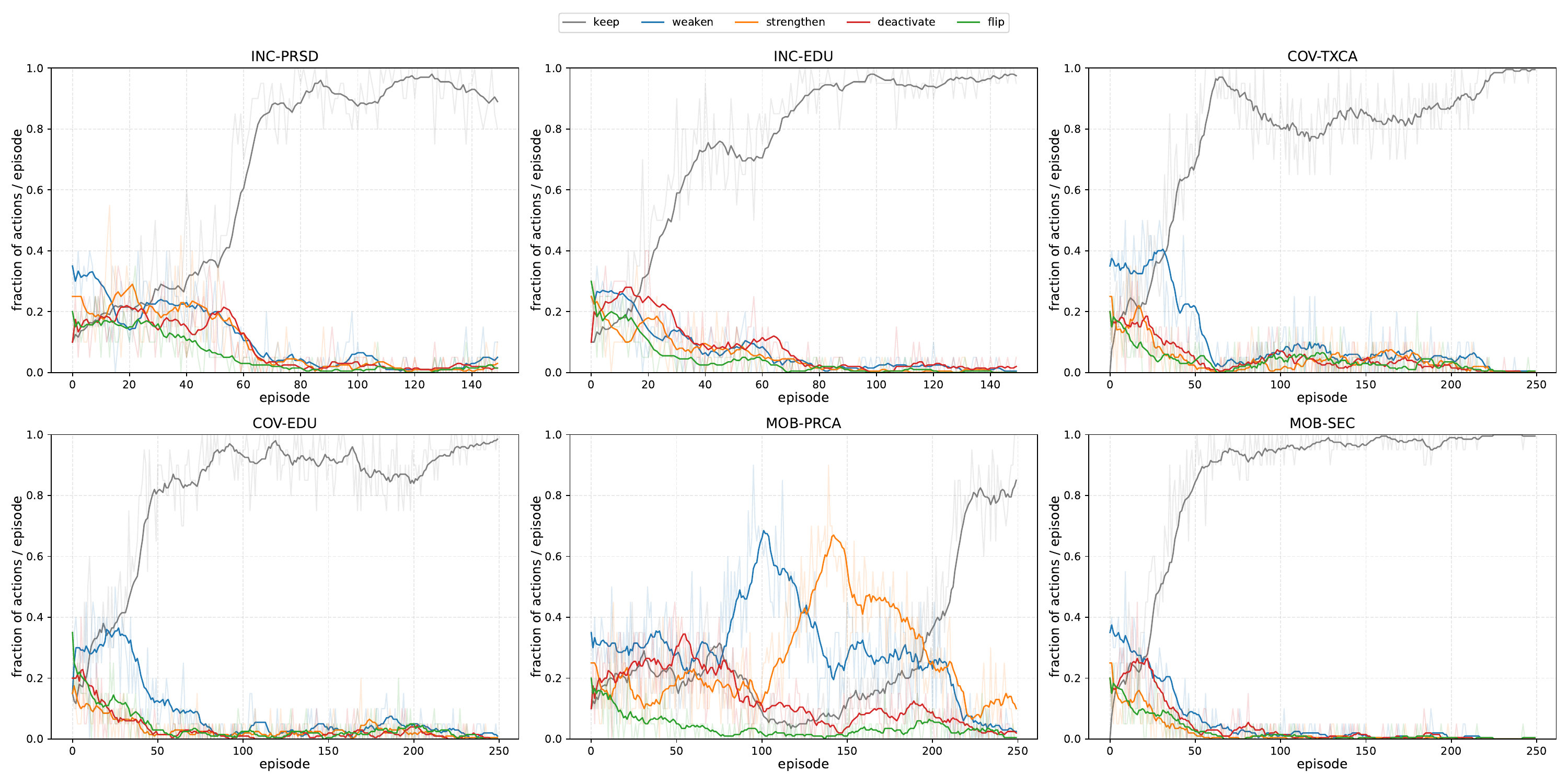}
\caption{Episode-wise edge-control action fractions for the six scenarios. Light traces show episode values, and dark traces show 10-episode moving averages. Each episode contains 20 actions.}
\label{fig:app-action-dynamics}
\end{figure*}

Figure~\ref{fig:app-action-dynamics} shows a common transition from exploratory edge adjustment to a keep-dominated policy. Over the final fifth of training, the mean keep fraction is 0.92 or higher in five scenarios and reaches 1.00 in \textsc{mob-sec}. The exception is \textsc{mob-prca}, where the final-fifth keep fraction is 0.73 and weaken and strengthen remain active at 0.08 and 0.14, respectively. This longer adjustment phase is consistent with a more heterogeneous policy search in that scenario. The rising keep fraction shows that later episodes make progressively fewer changes to the retained controls, consistent with the stable configurations evaluated in Appendix~\ref{app:full-results}.

\subsection{Episode-Return Dynamics}

\begin{figure*}[!t]
\centering
\includegraphics[width=0.90\textwidth]{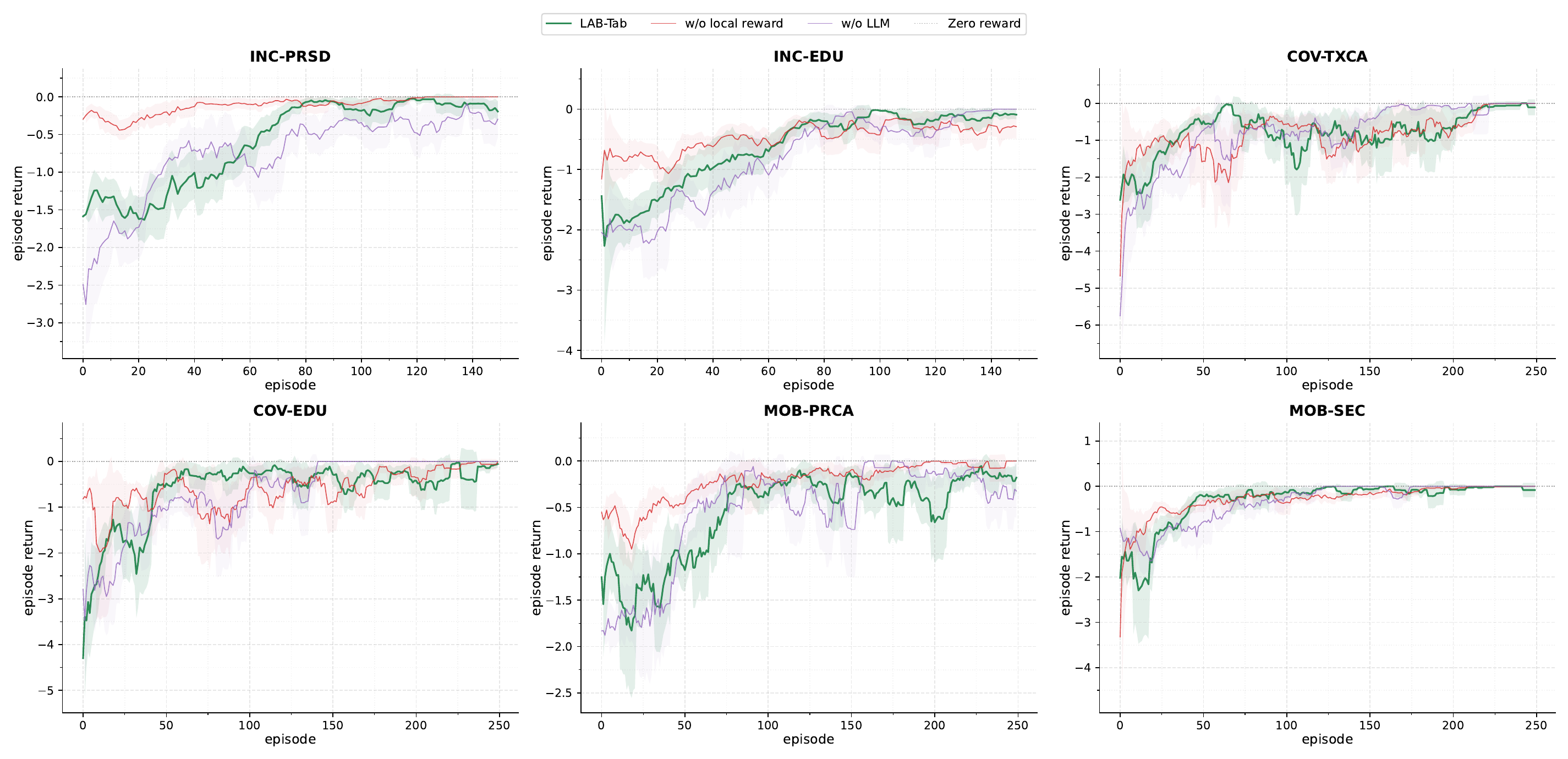}
\caption{Episode-return dynamics for LAB-Tab and two policy ablations. Solid curves are 10-episode rolling means, shaded regions are within-window 95\% smoothing bands, and the dotted line marks zero return.}
\label{fig:app-return-dynamics}
\end{figure*}

Let $r^{(e,k)}$ denote the reward at step $k$ of episode $e$. The episode return is
\begin{equation}
R_e=\sum_{k=1}^{20}r^{(e,k)}.
\end{equation}
Figure~\ref{fig:app-return-dynamics} reports this quantity for LAB-Tab, the variant without the local reward, and the variant without LLM edge proposals. The solid curves are 10-episode rolling means, and the shaded regions show pointwise 95\% smoothing bands computed within the same rolling window. Cross-seed variation is reported separately in the result tables.

The magnitude of the return decreases over training in all six scenarios, showing that later edge operations produce progressively smaller changes under each variant's objective. This behavior is most pronounced in \textsc{inc-prsd}, \textsc{cov-txca}, and \textsc{mob-sec}, where the rolling means approach zero near the end of training. The \textsc{mob-prca} traces remain more variable, consistent with the longer weaken--strengthen phase in Figure~\ref{fig:app-action-dynamics}. The variants use different reward definitions, and their final quality is summarized by UtilityGap, Coef--Cos, and Overall in the ablation results.

\FloatBarrier
\section{Complete Experimental Results}
\label{app:full-results}

All results use the six scenarios in Table~\ref{tab:app-scenarios}, five
reporting seeds, and the fixed Overall definition in
Appendix~\ref{app:metrics}. The 10\% headline and sensitivity entries use
the same runs.

\subsection{Scenario-Level Main Results}

\begin{table*}[!t]
\caption{Complete five-seed results for all six scenarios at the 10\%
target-data budget. Component entries are mean $\pm$ empirical standard
deviation; Overall entries are fixed-score means. Best means within each
scenario are bold.}
\label{tab:app-main-results}
\centering
\scriptsize
\setlength{\tabcolsep}{2.0pt}
\renewcommand{\arraystretch}{0.90}
\begin{tabular*}{\textwidth}{@{\extracolsep{\fill}}lccccc@{}}
\toprule
Method & JSD $\downarrow$ & WAPE $\downarrow$ & UGap $\downarrow$ & Coef--Cos $\uparrow$ & Overall $\downarrow$ \\
\midrule
\multicolumn{6}{l}{\textbf{\textsc{inc-prsd}}} \\
Gaussian Copula & $0.0318 \pm 0.0005$ & $0.5878 \pm 0.0091$ & $0.7089 \pm 0.0385$ & $0.1016 \pm 0.0084$ & $0.3200$ \\
MTabGen & $0.0402 \pm 0.0027$ & $0.4268 \pm 0.0055$ & $0.1567 \pm 0.0211$ & $\mathbf{0.7990 \pm 0.0032}$ & $0.1483$ \\
CTGAN & $0.0424 \pm 0.0057$ & $0.4642 \pm 0.0128$ & $0.2670 \pm 0.0408$ & $0.4177 \pm 0.0059$ & $0.2200$ \\
TVAE & $0.0456 \pm 0.0118$ & $0.4525 \pm 0.0134$ & $0.2383 \pm 0.0630$ & $0.7235 \pm 0.0072$ & $0.1770$ \\
CopulaGAN & $0.0451 \pm 0.0043$ & $0.4788 \pm 0.0074$ & $0.3409 \pm 0.0300$ & $0.5674 \pm 0.0066$ & $0.2149$ \\
SPADA & $0.0328 \pm 0.0001$ & $0.4374 \pm 0.0069$ & $0.1524 \pm 0.0256$ & $0.6296 \pm 0.0027$ & $0.1673$ \\
\textbf{LAB-Tab} & $\mathbf{0.0055 \pm 0.0002}$ & $\mathbf{0.4258 \pm 0.0067}$ & $\mathbf{0.1093 \pm 0.0171}$ & $0.7166 \pm 0.0024$ & $\mathbf{0.1367}$ \\
\midrule
\multicolumn{6}{l}{\textbf{\textsc{inc-edu}}} \\
Gaussian Copula & $0.0379 \pm 0.0003$ & $0.6026 \pm 0.0094$ & $1.1755 \pm 0.0505$ & $0.2508 \pm 0.0047$ & $0.3364$ \\
MTabGen & $0.0220 \pm 0.0029$ & $\mathbf{0.3622 \pm 0.0054}$ & $\mathbf{0.1014 \pm 0.0083}$ & $0.9244 \pm 0.0012$ & $\mathbf{0.1069}$ \\
CTGAN & $0.0421 \pm 0.0025$ & $0.4076 \pm 0.0122$ & $0.2450 \pm 0.0346$ & $0.8317 \pm 0.0035$ & $0.1578$ \\
TVAE & $0.0921 \pm 0.0086$ & $0.4724 \pm 0.0622$ & $0.5107 \pm 0.0512$ & $0.3832 \pm 0.0081$ & $0.2750$ \\
CopulaGAN & $0.0448 \pm 0.0030$ & $0.3701 \pm 0.0119$ & $0.1428 \pm 0.0361$ & $0.8710 \pm 0.0043$ & $0.1311$ \\
SPADA & $0.0264 \pm 0.0002$ & $0.3842 \pm 0.0047$ & $0.1704 \pm 0.0089$ & $\mathbf{0.9386 \pm 0.0007}$ & $0.1230$ \\
\textbf{LAB-Tab} & $\mathbf{0.0008 \pm 0.0004}$ & $0.3639 \pm 0.0065$ & $0.1425 \pm 0.0199$ & $0.8385 \pm 0.0008$ & $0.1184$ \\
\midrule
\multicolumn{6}{l}{\textbf{\textsc{cov-txca}}} \\
Gaussian Copula & $0.0054 \pm 0.0002$ & $0.1286 \pm 0.0031$ & $0.1877 \pm 0.0179$ & $0.9761 \pm 0.0019$ & $0.0729$ \\
MTabGen & $0.0100 \pm 0.0009$ & $0.1831 \pm 0.0116$ & $\mathbf{0.0705 \pm 0.0040}$ & $0.9833 \pm 0.0002$ & $0.0608$ \\
CTGAN & $0.0090 \pm 0.0012$ & $0.2002 \pm 0.0136$ & $0.1021 \pm 0.0149$ & $0.9315 \pm 0.0014$ & $0.0767$ \\
TVAE & $0.0538 \pm 0.0059$ & $0.3618 \pm 0.0268$ & $0.0948 \pm 0.0257$ & $\mathbf{0.9944 \pm 0.0007}$ & $0.1082$ \\
CopulaGAN & $0.0105 \pm 0.0012$ & $0.1948 \pm 0.0127$ & $0.0927 \pm 0.0110$ & $0.9807 \pm 0.0016$ & $0.0682$ \\
SPADA & $0.0026 \pm {<}0.0001$ & $0.0858 \pm 0.0013$ & $0.1863 \pm 0.0070$ & $-0.2254 \pm 0.0231$ & $0.2131$ \\
\textbf{LAB-Tab} & $\mathbf{0.0002 \pm 0.0001}$ & $\mathbf{0.0198 \pm 0.0020}$ & $0.0792 \pm 0.0171$ & $0.9918 \pm 0.0011$ & $\mathbf{0.0243}$ \\
\midrule
\multicolumn{6}{l}{\textbf{\textsc{cov-edu}}} \\
Gaussian Copula & $0.0207 \pm 0.0003$ & $0.2297 \pm 0.0027$ & $0.1202 \pm 0.0132$ & $0.9618 \pm 0.0026$ & $0.0858$ \\
MTabGen & $0.0220 \pm 0.0036$ & $0.2343 \pm 0.0369$ & $0.0046 \pm 0.0115$ & $0.9733 \pm 0.0006$ & $0.0599$ \\
CTGAN & $0.0253 \pm 0.0025$ & $0.2869 \pm 0.0210$ & $0.0708 \pm 0.0108$ & $0.9618 \pm 0.0024$ & $0.0862$ \\
TVAE & $0.0426 \pm 0.0038$ & $0.3052 \pm 0.0050$ & $0.0211 \pm 0.0332$ & $0.9845 \pm 0.0007$ & $0.0809$ \\
CopulaGAN & $0.0285 \pm 0.0022$ & $0.2946 \pm 0.0310$ & $\mathbf{0.0012 \pm 0.0127}$ & $\mathbf{0.9853 \pm 0.0013}$ & $0.0693$ \\
SPADA & $0.0179 \pm 0.0001$ & $0.2079 \pm 0.0014$ & $0.1336 \pm 0.0186$ & $-0.0132 \pm 0.0264$ & $0.2056$ \\
\textbf{LAB-Tab} & $\mathbf{0.0006 \pm 0.0002}$ & $\mathbf{0.0479 \pm 0.0086}$ & $0.0048 \pm 0.0083$ & $0.7590 \pm 0.0078$ & $\mathbf{0.0430}$ \\
\midrule
\multicolumn{6}{l}{\textbf{\textsc{mob-prca}}} \\
Gaussian Copula & $0.0161 \pm 0.0004$ & $0.1720 \pm 0.0036$ & $0.1043 \pm 0.0124$ & $0.6518 \pm 0.0472$ & $0.1096$ \\
MTabGen & $0.0241 \pm 0.0013$ & $0.2587 \pm 0.0127$ & $\mathbf{0.0507 \pm 0.0072}$ & $-0.5180 \pm 0.0246$ & $0.2619$ \\
CTGAN & $0.0288 \pm 0.0014$ & $0.2915 \pm 0.0194$ & $0.1046 \pm 0.0098$ & $\mathbf{0.6697 \pm 0.0431}$ & $0.1318$ \\
TVAE & $0.0349 \pm 0.0077$ & $0.2950 \pm 0.0203$ & $0.0655 \pm 0.0145$ & $0.2847 \pm 0.0442$ & $0.1743$ \\
CopulaGAN & $0.0362 \pm 0.0017$ & $0.3187 \pm 0.0244$ & $0.0755 \pm 0.0110$ & $-0.7473 \pm 0.0338$ & $0.3094$ \\
SPADA & $0.0158 \pm 0.0003$ & $0.1624 \pm 0.0020$ & $0.0946 \pm 0.0107$ & $0.5833 \pm 0.0390$ & $0.1143$ \\
\textbf{LAB-Tab} & $\mathbf{0.0004 \pm 0.0001}$ & $\mathbf{0.0319 \pm 0.0048}$ & $0.0643 \pm 0.0049$ & $0.5140 \pm 0.0092$ & $\mathbf{0.0837}$ \\
\midrule
\multicolumn{6}{l}{\textbf{\textsc{mob-sec}}} \\
Gaussian Copula & $0.0049 \pm 0.0002$ & $0.1157 \pm 0.0019$ & $0.0948 \pm 0.0278$ & $-0.6725 \pm 0.0418$ & $0.2584$ \\
MTabGen & $0.0093 \pm 0.0009$ & $0.1870 \pm 0.0131$ & $0.0668 \pm 0.0157$ & $-0.3176 \pm 0.0243$ & $0.2231$ \\
CTGAN & $0.0117 \pm 0.0030$ & $0.2081 \pm 0.0328$ & $0.0852 \pm 0.0240$ & $0.2771 \pm 0.0302$ & $0.1573$ \\
TVAE & $0.0263 \pm 0.0096$ & $0.2540 \pm 0.0438$ & $\mathbf{0.0029 \pm 0.0258}$ & $\mathbf{0.7809 \pm 0.0191}$ & $\mathbf{0.0882}$ \\
CopulaGAN & $0.0339 \pm 0.0041$ & $0.2982 \pm 0.0318$ & $0.0579 \pm 0.0254$ & $0.7151 \pm 0.0399$ & $0.1189$ \\
SPADA & $0.0047 \pm 0.0001$ & $0.1123 \pm 0.0009$ & $0.0912 \pm 0.0226$ & $-0.2649 \pm 0.0186$ & $0.2059$ \\
\textbf{LAB-Tab} & $\mathbf{0.0018 \pm 0.0002}$ & $\mathbf{0.0707 \pm 0.0082}$ & $0.0367 \pm 0.0159$ & $0.0587 \pm 0.0046$ & $0.1437$ \\
\bottomrule
\end{tabular*}
\end{table*}

\raggedbottom
\subsection{Results across Target-Data Budgets}

Table~\ref{tab:app-budget-anchors} summarizes the main target-budget anchors.
LAB-Tab leads seven of eleven budgets, with its largest gains in the low-data
regime. At 1\%, it reduces Overall by 54.6\% relative to CTGAN. MTabGen leads
at 40\%, 60\%, 90\%, and 100\%, and the two methods remain close at the
full-data endpoint. This crossover is consistent with source-informed
adaptation being most useful when target evidence is scarce.

\medskip
\noindent\begin{minipage}{\columnwidth}
\centering
\captionof{table}{Selected anchors from the target-data sensitivity analysis.}
\label{tab:app-budget-anchors}
\footnotesize
\setlength{\tabcolsep}{4pt}
\renewcommand{\arraystretch}{1.04}
\begin{tabular}{@{}lrr@{}}
\toprule
Setting & LAB-Tab & Reference \\
\midrule
1\% Overall   & \textbf{0.1499} & CTGAN: 0.3299 \\
10\% Overall  & \textbf{0.0916} & CTGAN: 0.1383 \\
100\% Overall & 0.1751 & \textbf{MTabGen: 0.1650} \\
Budget wins   & \textbf{7/11} & MTabGen: 4/11 \\
\bottomrule
\end{tabular}
\end{minipage}

\subsection{Ablation Summary}

Table~\ref{tab:app-ablation-summary} reports the fixed-scale ablation
averages. The results show complementary roles across the components. LLM
augmentation mainly improves conditional and feature--label structure, PPO
coordinates inherited and added edges, and local feedback improves the
balance of utility and association preservation.

\medskip
\noindent\begin{minipage}{\columnwidth}
\centering
\captionof{table}{Ablation results averaged over five seeds and six scenarios.}
\label{tab:app-ablation-summary}
\footnotesize
\setlength{\tabcolsep}{2.0pt}
\renewcommand{\arraystretch}{1.02}
\begin{tabular}{@{}lrrrrr@{}}
\toprule
Method & JSD$\downarrow$ & WAPE$\downarrow$ & UGap$\downarrow$ & Coef.$\uparrow$ & Overall$\downarrow$ \\
\midrule
Source BN       & 0.0318 & 0.3137 & 0.1861 & 0.5280 & 0.1633 \\
Target BN       & \textbf{0.0011} & 0.1815 & 0.1550 & 0.4137 & 0.1382 \\
Source BN + MAP & \textbf{0.0011} & 0.1641 & 0.0923 & 0.6082 & 0.1003 \\
\midrule
w/o LLM         & 0.0015 & 0.1659 & 0.0929 & 0.4478 & 0.1211 \\
w/o PPO         & 0.0013 & 0.1626 & 0.0766 & 0.5803 & 0.0997 \\
Greedy edits    & 0.0015 & 0.1707 & 0.0995 & 0.5536 & 0.1087 \\
w/o local reward& 0.0016 & \textbf{0.1584} & 0.0865 & 0.4892 & 0.1137 \\
\midrule
\textbf{LAB-Tab}& 0.0016 & 0.1600 & \textbf{0.0728} & \textbf{0.6464} & \textbf{0.0916} \\
\bottomrule
\end{tabular}
\end{minipage}